\documentclass[letterpaper]{article} 
\usepackage{aaai24}  
\usepackage{times}  
\usepackage{helvet}  
\usepackage{courier}  
\usepackage[hyphens]{url}  
\usepackage{graphicx} 
\usepackage{natbib}  
\usepackage{caption} 
\usepackage{algorithm}
\usepackage{algorithmic}

\usepackage{newfloat}
\usepackage{listings}

\usepackage{multirow}
\usepackage{amsmath}
\usepackage{booktabs}
\usepackage{pifont}
\usepackage{subcaption}
\usepackage{fdsymbol}
\usepackage{bm}
\usepackage{marvosym}
\usepackage{epsfig}
\usepackage{graphicx}
\usepackage{booktabs}
\usepackage{hhline}
\DeclareCaptionStyle{ruled}{labelfont=normalfont,labelsep=colon,strut=off} 
\floatstyle{ruled}
\newfloat{listing}{tb}{lst}{}
\floatname{listing}{Listing}
\title{Cross-Covariate Gait Recognition: A Benchmark}
\author {
    Shinan Zou\textsuperscript{\rm 1},
    Chao Fan\textsuperscript{\rm 2,3},
    Jianbo Xiong\textsuperscript{\rm 1},
    Chuanfu Shen\textsuperscript{\rm 2, 4},
    Shiqi Yu\textsuperscript{\rm 2,3},
    Jin Tang\textsuperscript{\rm 1\thanks{Corresponding Author}}
}
\affiliations {
    \textsuperscript{\rm 1}School of Automation, Central South University\\
    \textsuperscript{\rm 2}Department of Computer Science and Engineering, Southern University of Science and Technology\\
    \textsuperscript{\rm 3}Research Institute of Trustworthy Autonomous System, Southern University of Science and Technology\\
    \textsuperscript{\rm 4}The University of Hong Kong\\
\{zoushinan, jianbo\_x, tjin\}@csu.edu.cn, \{12131100, 11950016\}@mail.sustech.edu.cn, yusq@sustech.edu.cn
}
\usepackage{bibentry}

\begin{document}

\maketitle

\begin{abstract}
     Gait datasets are essential for gait research. However, this paper observes that present benchmarks, whether conventional constrained or emerging real-world datasets, fall short regarding covariate diversity. To bridge this gap, we undertake an arduous 20-month effort to collect a cross-covariate gait recognition (CCGR) dataset. The CCGR dataset has 970 subjects and about 1.6 million sequences; almost every subject has 33 views and 53 different covariates. Compared to existing datasets, CCGR has both population and individual-level diversity. In addition, the views and covariates are well labeled, enabling the analysis of the effects of different factors. CCGR provides multiple types of gait data, including RGB, parsing, silhouette, and pose, offering researchers a comprehensive resource for exploration. In order to delve deeper into addressing cross-covariate gait recognition, we propose parsing-based gait recognition (ParsingGait) by utilizing the newly proposed parsing data. We have conducted extensive experiments. Our main results show: 1) Cross-covariate emerges as a pivotal challenge for practical applications of gait recognition. 2) ParsingGait demonstrates remarkable potential for further advancement. 3) Alarmingly, existing SOTA methods achieve less than 43\% accuracy on the CCGR, highlighting the urgency of exploring cross-covariate gait recognition. Link: \url{https://github.com/ShinanZou/CCGR}.
 
	\end{abstract}
	
    \section{Introduction}
    \label{sec:intro}
	
    Gait recognition aims to use physiological and behavioral characteristics extracted from walking videos to certify individuals' identities. Compared to other biometric modalities, such as face, fingerprints, and iris, gait patterns have the distinct advantage of being extracted from a distance in uncontrolled environments.
    These strengths place gait recognition as an effective solution for security applications. 

\begin{figure}[t]
		\centering
		\includegraphics[width=1\linewidth] {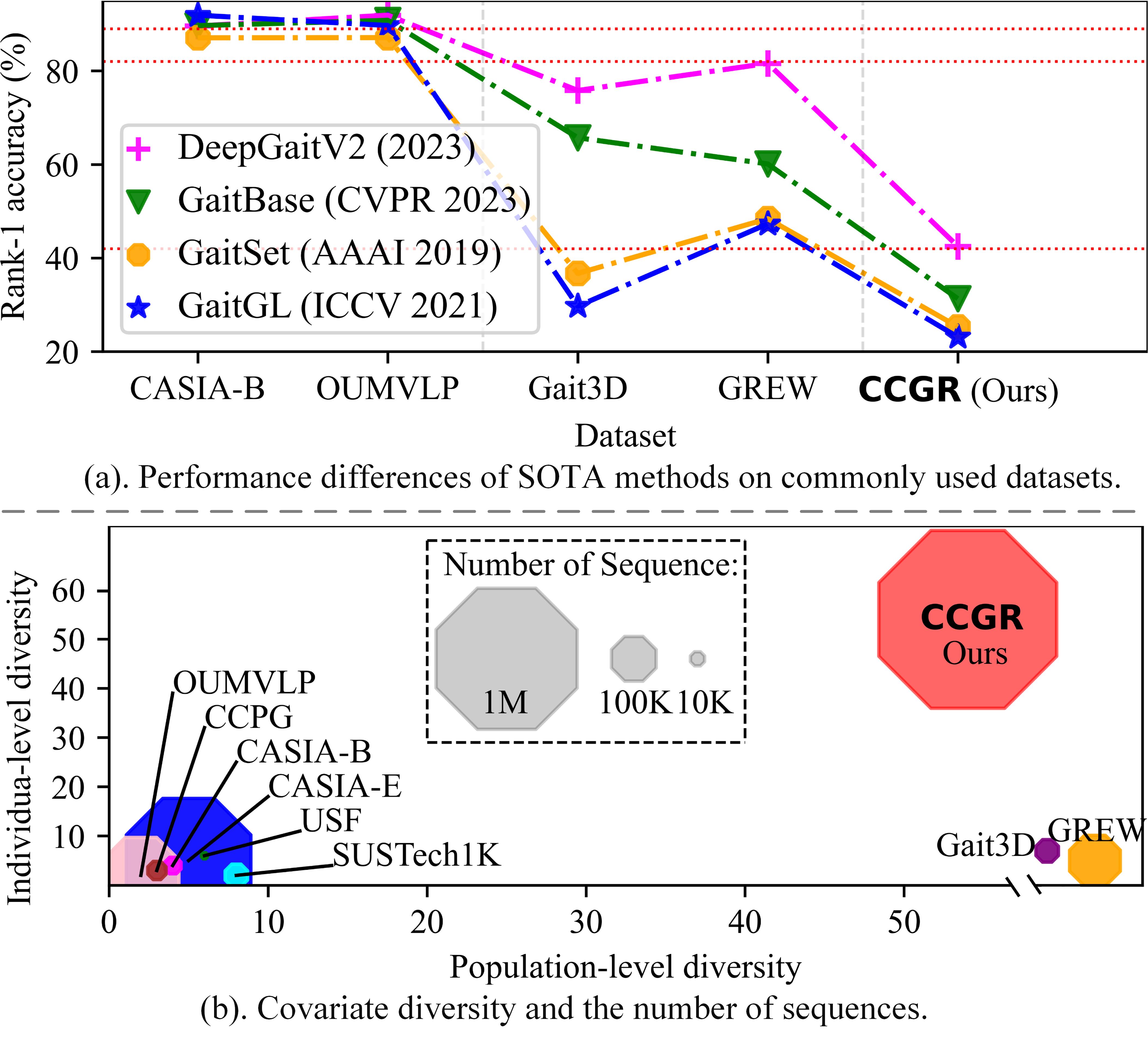}
		\caption{Differences between CCGR and other datasets. 
  Population-level diversity is roughly quantified by the count of covariate categories present within the whole dataset. 
  Correspondingly, individual-level diversity is measured by the count of covariate categories for each subject. 
  Here, the population-level diversity of Gait3D and GREW is rich, but the exact amount is unknown due to the wild scenarios. 
  } 
		\label{figs1}
   \end{figure}

    In the latest literature, the research on gait recognition is developing rapidly, with the evaluation benchmark developing from early indoor to outdoor environments. During this remarkable journey, most representative gait models~\cite{9, 68} boasting historical progress have unexpectedly performed unsatisfactory results when faced with emerging challenges posed by real-world gait datasets such as GREW~\cite{76} and Gait3D~\cite{75}. Surprisingly, successive works~\cite{124, 126} quickly address this performance gap to a large extent, rekindling the promise of gait recognition for practical applications, as illustrated in Figure~\ref{figs1}(a). However, this paper argues that the gait recognition task is much more challenging than these datasets have defined.
	
	
   \begin{table*}[t]
		\centering
		\caption{Comparison of CCGR with existing datasets. Sil., Inf., A., and 3DM. mean silhouette, infrared, audio, and 3D Mesh\&SMPL.  \#Id, \#Seq, and \#Cam refer to the number of identities, sequences, and cameras. BAC, CO, GR, BR, DU, IN, BA, TR, SH, CL, UB, OC, NI, and WS  are abbreviations of backpack, concrete, grass,  briefcase, duration, incline, ball, treadmill, shoes, clothing, umbrella, uniform, occlusion, night and walking style. CMU MoBo \cite{93}; SOTON \cite{95}; USF \cite{96}; CASIA-B \cite{25}; CASIA-C \cite{97}; OU-ISIR Speed \cite{98}; OU-ISIR Cloth \cite{99}; OU-ISIR MV \cite{100}; OU-LP \cite{101}; TUM GAID \cite{103};  OU-LP Age \cite{104}; OU-MVLP \cite{26,107,108}; OU-LP Bag \cite{105}; GREW \cite{76}; ReSGait \cite{77}; UAV-Gait \cite{119};  Gait3D \cite{75}; CASIA-E \cite{127}; CCPG \cite{129}; SUSTech1K \cite{128}. \footnotesize} 
		\resizebox{2.1\columnwidth}{!}{
			\begin{tabular}{l|c|c|c|c|c|c|c}
				\toprule[1pt]
				Dataset               & \#Id          & \#Seq             & \#Cam                & Data types                                                                                & Covariates except view                  & Environment  &  Diversity       \\ \hline 
				CMU MoBo              & 25             & 600                & 6                  & RGB, Sil.                                                                        & TR, Speed, BA, IN                                                                               & Controlled  & Not Rich        \\
				SOTON                  & 115            & 2,128              & 2                & RGB, Sil.                                                                                 & TR                                                                                              & Controlled     & Not Rich     \\
				USF                   & 122            & 1,870              & 2                 & RGB                                                                                       & CO, GR, SH, BR, DU                                                                              & Controlled        & Not Rich  \\
				CASIA-B               & 124            & 13,640             & 11                & RGB, Sil.                                                                                 & Coat, Bag                                                                                       & Controlled       & Not Rich   \\
				CASIA-C               & 153            & 1,530              & 1                 & Inf., Sil.                                                                                & SP, Bag                                                                                         & Controlled      & Not Rich    \\
				OU-ISIR Speed         & 34             & 612                & 1                  & Sil.                                                                                      & TR, Speed                                                                                       & Controlled     & Not Rich     \\
				OU-ISIR Cloth        & 68             & 2,764              & 1                & Sil.                                                                                      & TR, CL                                                                                    & Controlled        & Not Rich  \\
				OU-ISIR MV            & 168            & 4,200              & 25                 & Sil.                                                                                      & TR                                                                                              & Controlled      & Not Rich    \\
				OU-LP              & 4,007          & 7,842              & 2                  & Sil.                                                                                      & None                                                                                            & Controlled       & Not Rich   \\
				TUM GAID            & 305            & 3,370              & 1                  & RGB, Depth, A.                                                                            & DU, BAC, SH                                                                                     & Controlled       & Not Rich   \\
				OU-LP Age            & 63,846         & 63,846             & 1                 & Sil.                                                                                      & Age                                                                                             & Controlled      & Not Rich    \\
				OU-MVLP              & 10,307         & 288,596            & 14                & Sil., Pose, 3DM.                                                                                      & None                                                                                            & Controlled      & Not Rich    \\
				OU-LP Bag             & 62,528         & 187,584            & 1                 & Sil.                                                                                      & Carrying                                                                                        & Controlled       & Not Rich   \\
				GREW                  & 26,345         & 128,671            & 882            & Sil., Flow, Pose                                                                          & Free walking                                                                                    & Wild             & Population-Level   \\
				ReSGait               & 172            & 870                & 1                  & Sil., Pose                                                                                & Free walking                                                                                    & Wild           & Population-Level     \\
				UAV-Gait            & 202         & 9,895             & 6               & Sil., pose                                                                                      & None                                                                                             & Controlled       & Not Rich   \\
				Gait3D                & 4,000          & 25,309             & 39            & Sil., Pose, 3DM.                                                                          & Free walking                                                                                    & Wild           & Population-Level     \\ 
				CASIA-E                & 1,014          & 778,752             & 26              & RGB, Sil.                                                                          & Bag, CL, WS                                                                                     & Controlled   & Not Rich  \\
				CCPG                & 200          & 16,566             & 10             & RGB, Sil.                                                                       &  CL                                                                                    & Controlled        & Not Rich    \\
				SUSTech1K                & 1,050          & 25,279           & 12      & RGB, Sil., 3DP                                                                          & Bag, CL, UB, OC, NI                                                                                    & Controlled         & Not Rich   \\ \hline
				\textbf{CCGR (ours)}         & 970 & \textbf{1,580,617} & 33  & \begin{tabular}[c]{@{}c@{}}RGB, \textbf{Parsing}, \\ Sil., Pose\end{tabular} & \textbf{\begin{tabular}[c]{@{}c@{}}53 types per subject,\\  as detailed in Figure \ref{Fig3}.\end{tabular}} & Controlled & \textbf{\begin{tabular}[c]{@{}c@{}}Population- and \\ Individual-Level \end{tabular}} \\ \bottomrule[1pt]
			\end{tabular}
		}
		\label{Tab1}
    \end{table*}

    In general, previous indoor gait datasets often require subjects repeatedly walk along fixed paths while introducing variations in clothing and carrying. This approach yields controllable and well-annotated data, facilitating the early exploration of key covariates influencing recognition accuracy. However, as shown in Fig.~\ref{figs1}(b), these datasets fall short regarding \textbf{population-level diversity}, as subjects of them contain the same limited group of covariates. Conversely, the emergence of outdoor datasets effectively addresses this limitation due to their real-world collection scenarios. Although their data distribution closely mirrors practical applications, we contend that current outdoor gait datasets lack \textbf{individual-level diversity}, as each subject typically contributes no more than seven variants (sequences) on average. This situation gives rise to two potential drawbacks for research: a) A majority of data pairs may qualify as ``easy cases" owing to limited collection areas and short-term data gathering. b) The lack of fine annotations blocks exploring critical challenges relevant to real-world applications. More details of the existing dataset are in Table \ref{Tab1}.

    \begin{figure*}[t]
		\centering

		\includegraphics[width=1\linewidth]{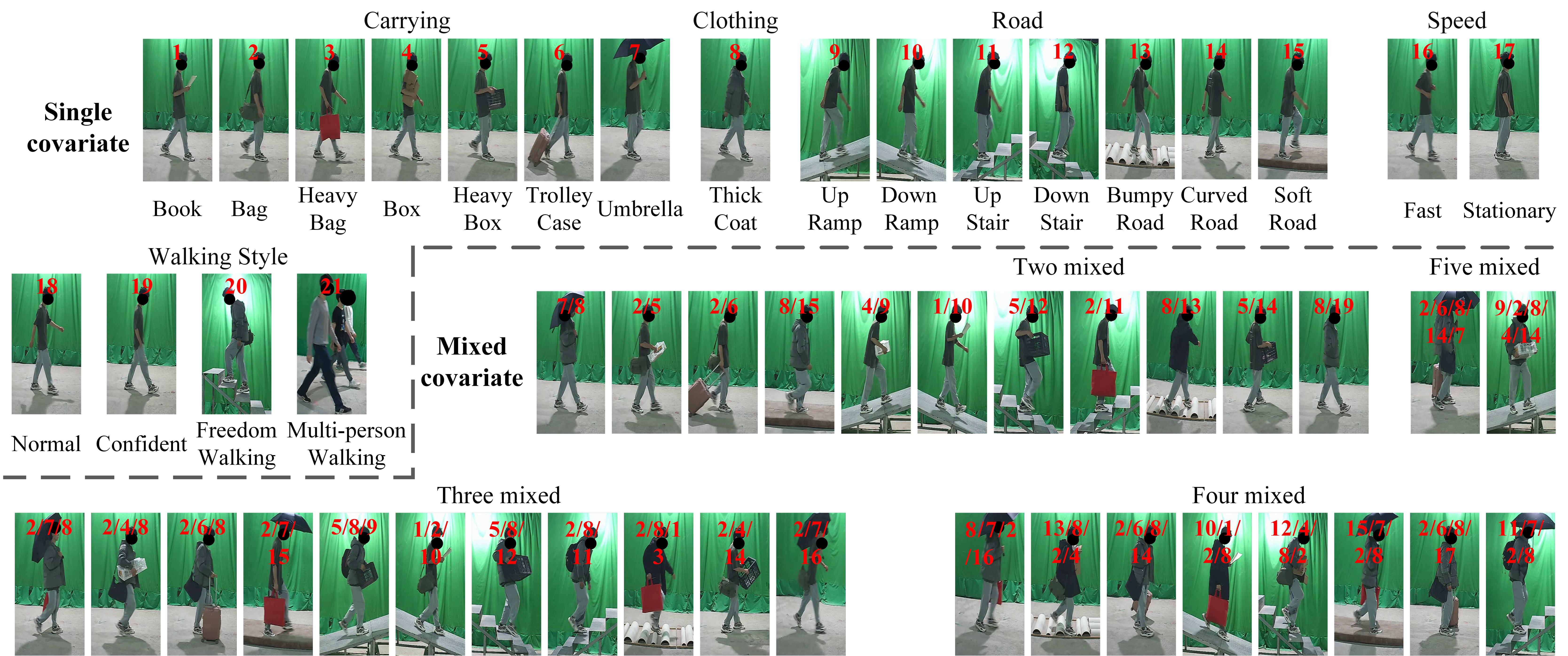}
		\caption{Examples of 53 covariates in CCGR. For a single covariate (the 1st row and the left of the 2nd row), the red numbers at the top of the pictures are indices of the covariates. For mixed covariates, numbers separated by ``/" at the top of the picture indicate the co-occur of multi-single covariates corresponding to these numbers.\footnotesize}
		\label{Fig3}
	
    \end{figure*}
    To overcome these limitations, we propose a novel gait recognition benchmark that introduces both population-level and individual-level diversity, named \textbf{Cross-Covariate Gait Recognition} or \textbf{CCGR}. Statistically, the CCGR dataset covers 970 subjects and approximately 1.6 million walking sequences. These sequences span 53 distinct walking conditions and 33 different filming views. Thus, each subject within CCGR ideally contains a comprehensive collection of $53\times33=1,749$ sequences. Notably, the walking conditions are widely distributed and well annotated, encompassing diverse factors such as carried items (book, bag, box, umbrella, trolley case, heavy bag, and heavy box), road types (up the stair, down the stair, up the ramp, down the ramp, bumpy road, soft road, and curved road), styles of walking (fast, stationary, normal, hands in pockets, free, and crowd), and more. The all-side camera array consisting of 33 cameras is installed at five different heights, effectively simulating the pitching angles of typical CCTVs. Every subject is recruited through a transparent process and accompanied by written consent. The age range of subjects spans from 6 to 70 years. The dataset encompasses raw RGB sequences. Releasing RGB images can facilitate the exploration of camera-based gait representations, and this paper officially provides common gait data like silhouette, parsing, and pose. CCGR will be made publicly available for research purposes.

    Equipped with the proposed CCGR, we re-implement several representative state-of-the-art methods and investigate that: 1) Cross-covariate gait recognition is more challenging than that simulated by previous gait datasets, as the achieved best rank-1 accuracy is only $42.5\%$. 2) Certain less-researched covariates, such as the crowd, umbrella, overhead view, walking speed,  road, mixed covariate, and more, significantly degrade the recognition accuracy. 3) The more covariates involved, regardless of population-level and individual-level diversity perspectives, the more challenging gait recognition becomes.  

    To solve complex covariate problems, this paper further introduces human parsing, which contains many semantic characteristics that describe body parts, to form a parsing-based baseline framework termed \textbf{ParsingGait}. In practice, we instantiate the backbone of PrasingGait using various silhouette-based gait models, consistently achieving significant enhancements. By this means, this paper highlights the value of informative gait representations like human parsing images for gait pattern description.  


	In summary, our main contributions are as follows:
	
	$\bullet$ We present the first well-annotated million-sequence-level gait recognition benchmark called CCGR, designed to research cross-covariate gait recognition deeply.

	$\bullet$ We propose an efficient, compatible, and feasible parsing-based baseline framework named ParsingGait.
	
	$\bullet$ We begin by evaluating existing algorithms to establish a baseline, then validating the effectiveness of ParsingGait. Next, we demonstrate the necessity of incorporating both 
 population- and individual-level diversity. Finally, we thoroughly explore the impact of covariates and views.


    \section{The CCGR Benchmark}
    \label{sec:CCGR}
  

    \subsection{Covariates of CCGR}
	The dataset has 53 covariates; 21 are single covariates, while the remaining 32 are mixed covariates. Examples of the 53 covariates are shown in Figure \ref{Fig3}.  
	
	\textbf{Carrying:} We have defined seven carrying covariates: \textbf{book}, \textbf{bag}, \textbf{heavy bag}, \textbf{box}, \textbf{heavy box}, and \textbf{trolley case}, \textbf{umbrella}. We have prepared 12 different types for the bag category, including single-shoulder bags, double-shoulder bags, satchels, backpacks, and handbags. Similarly, we have prepared eight boxes with varying shapes and volumes for the box category. As for the trolley case, we have prepared options in both 20-inch and 28-inch sizes. When subjects are asked to carry a bag, box, or trolley case, they can choose from the props we have provided. In the case of the heavy bag and box, we have placed counterweights inside them, ranging from 8kg to 15kg, to simulate the desired weight.  
	
	\textbf{Clothing:} Regarding the \textbf{thick coat} covariates, we have prepared a selection of 20 clothing items, which include down coats, overcoats, windbreakers, jackets, and cotton coats. When subjects are instructed to wear a thick coat, they can choose from our clothing collection.
	
	\textbf{Road:} In addition to the normal road, we have prepared seven road covariates: \textbf{up/down the stair}, \textbf{up/down the ramp}, \textbf{bumpy road}, \textbf{soft (muddy) road}, and \textbf{curved road}. Ramps have a slope of $15^o$. Curved road means subjects are asked to walk a curved track instead of a straight path.

    \begin{figure*}[t]
		\centering

		\includegraphics[width=1\linewidth]{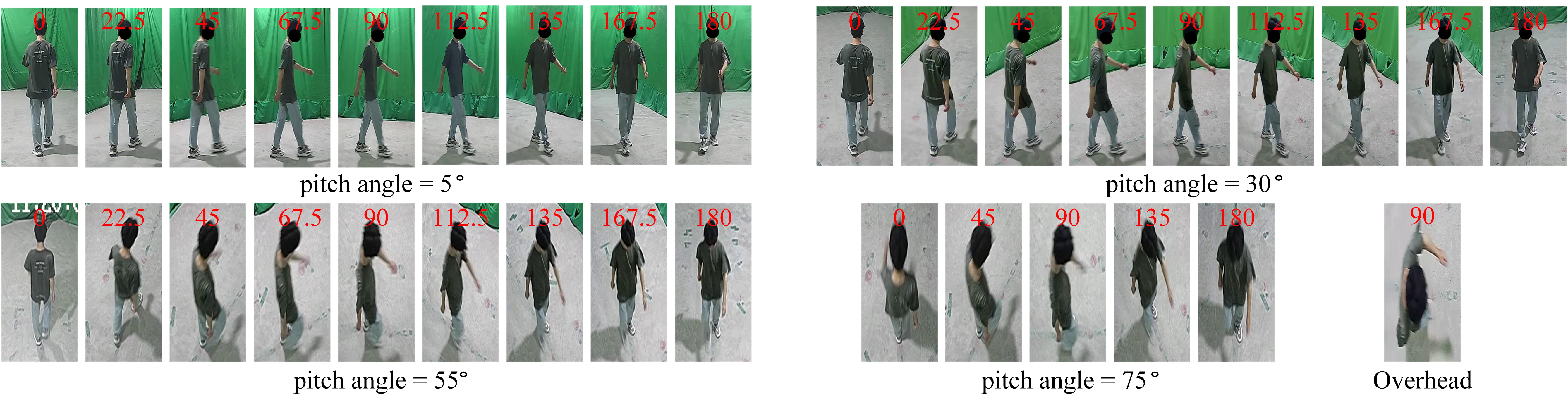}
		\caption{Examples of 33 views in CCGR. The red numbers at the top of the picture represent the horizontal angle.\footnotesize}
	
		\label{figs3}
    \end{figure*}
    \begin{figure}[t]
		\centering  
		\includegraphics[width=1\linewidth]{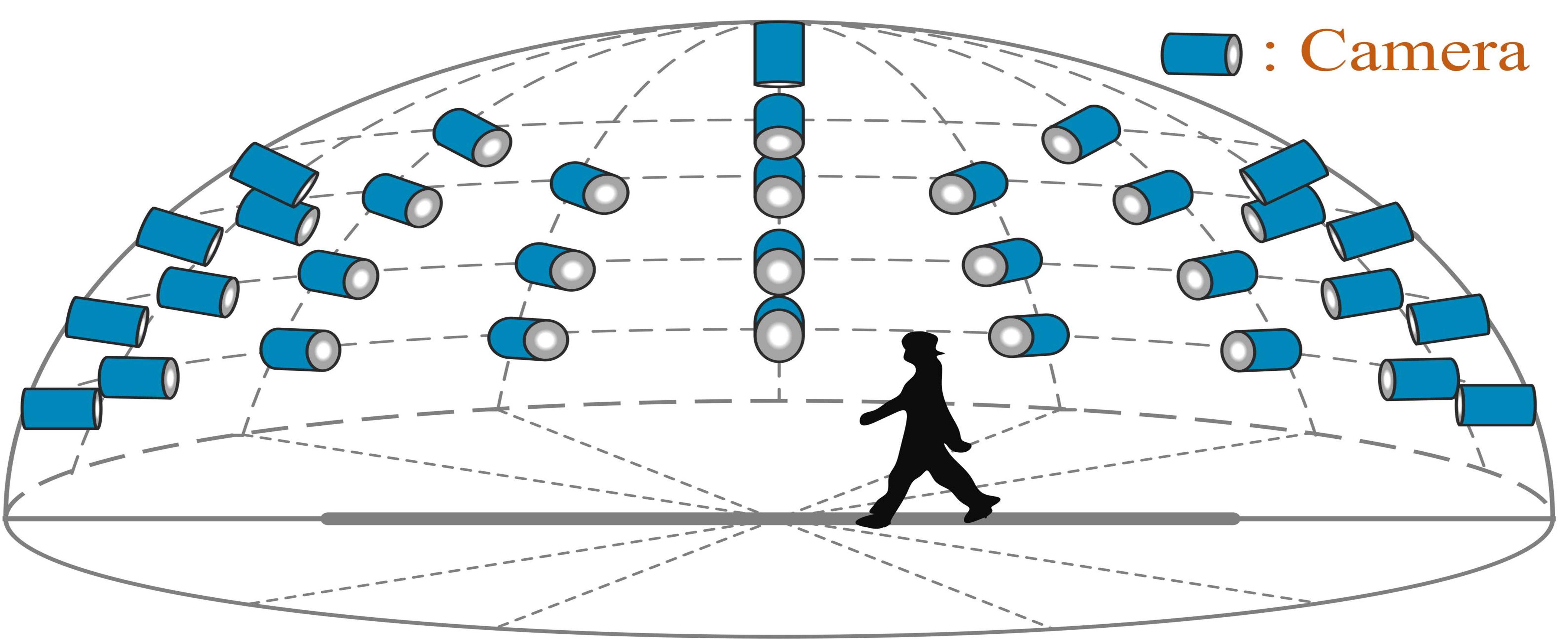}
		\caption{Camera setup in CCGR.\footnotesize}
		\label{figs2}
    \end{figure}
	\textbf{Speed:} In addition to the normal walking speed, we discuss two additional walking speeds: \textbf{fast} and \textbf{stationary}. Fast entails the subject walking at a speed close to a trot, while stationary refers to the subject remaining unmoving.
	
	\textbf{Walking Style:} The remaining four single covariates include \textbf{normal walking}, \textbf{confident}, \textbf{multi-person walking}, and \textbf{freedom walking}. Normal walking indicates walking on a horizontal path at a normal speed without wearing a thick coat or carrying any items. Confident means that subjects place their hands inside their pant or clothing pockets. Multi-person walking means multiple subjects walking together. Freedom walking means subjects are free to choose their carrying, clothing, road, and speed.
	
	\textbf{Mixed covariates:} In the real world, multiple covariates often co-occur. For instance, a man may wear a thick coat, carry a bag, and walk up a ramp. To simplify matters, we utilize mixed covariates to represent the co-occurrence of multiple covariates. In CCGR, we have designed 32 \textbf{mixed covariates} that are frequently encountered in daily life. Refer to Figure \ref{Fig3} for further details about these mixed covariates.
 

    \subsection{Views of CCGR}
    We rent a 500-square-meter warehouse and set up 33 cameras to collect data. Camera settings are shown in Figures \ref{figs2}. The cameras are divided into five layers, from bottom to top. Layer 5 is the overhead camera with a pitch angle of $90^o$. For the other four layers, the pitch angles from bottom to top are $5^o$, $30^o$, $55^o$, and $75^o$, and the horizontal angles of each layer increase from $0^o$ to $180^o$ counterclockwise. The frame size of the video files is $1280 \times 720$, and the frame rate is 25 fps. Figure \ref{figs3} shows the example with various views.
 

    \subsection{Extraction of Multiple Gait Data}
\begin{figure}[t]
    	\centering

    	\includegraphics[width=1\linewidth]{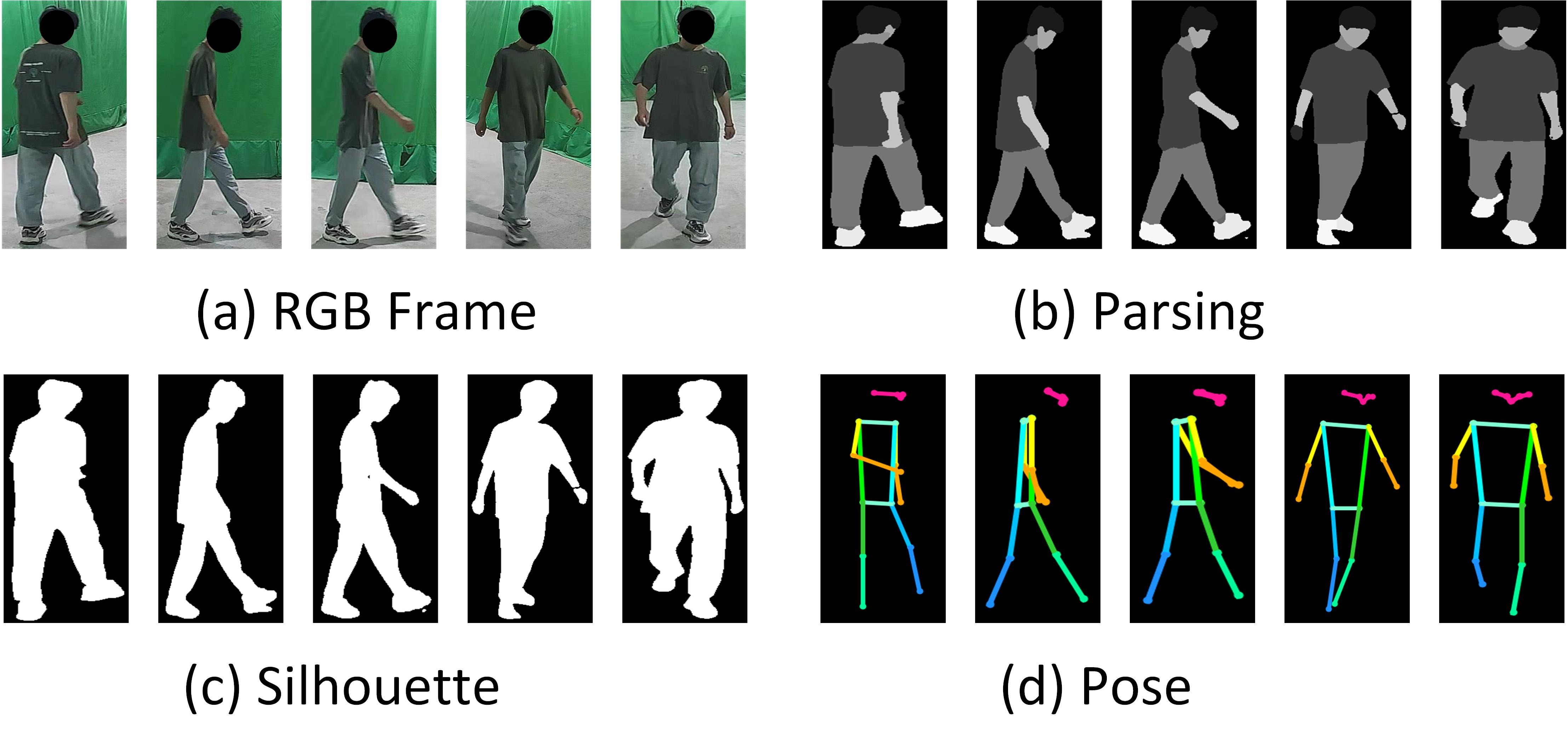}
    	\caption{Examples of different gait data in CCGR .\footnotesize}
    	\label{figs4}
    \end{figure}

     We offer various types of gait data, including RGB, parsing, silhouette, and pose; examples can be seen in Figure \ref{figs4}.
    
    \textbf{Parsing:} Predicting the semantic category of each pixel on the human body is a fundamental task in computer vision, often referred to as human parsing \cite{78,79,80,81}. We uses QANet \cite{87} for parsing extraction. QANet takes an RGB image as its input and produces the semantic category of each pixel on the human body, including hair, face, and left leg. Initially, QANet employed integers ranging from 0 to 19 to represent these different categories. To facilitate visualization and image pruning, we multiply these integers by 13 to generate a grayscale image.
    
    
    \textbf{Silhouette:} 
    We generate the silhouettes by directly binarizing the previously acquired parsing images. 
    We have also tried the instance and semantic segmentation algorithms but attained relatively inferior gait recognition accuracy.

    \textbf{Pose:} We use HRNet \cite{111} to extract 2D Pose. We also try AlphaPose \cite{64} and Openpose \cite{63}, which result in inferior accuracy.
    
    \begin{figure}[t]
		\centering
		\includegraphics[width=1\linewidth]{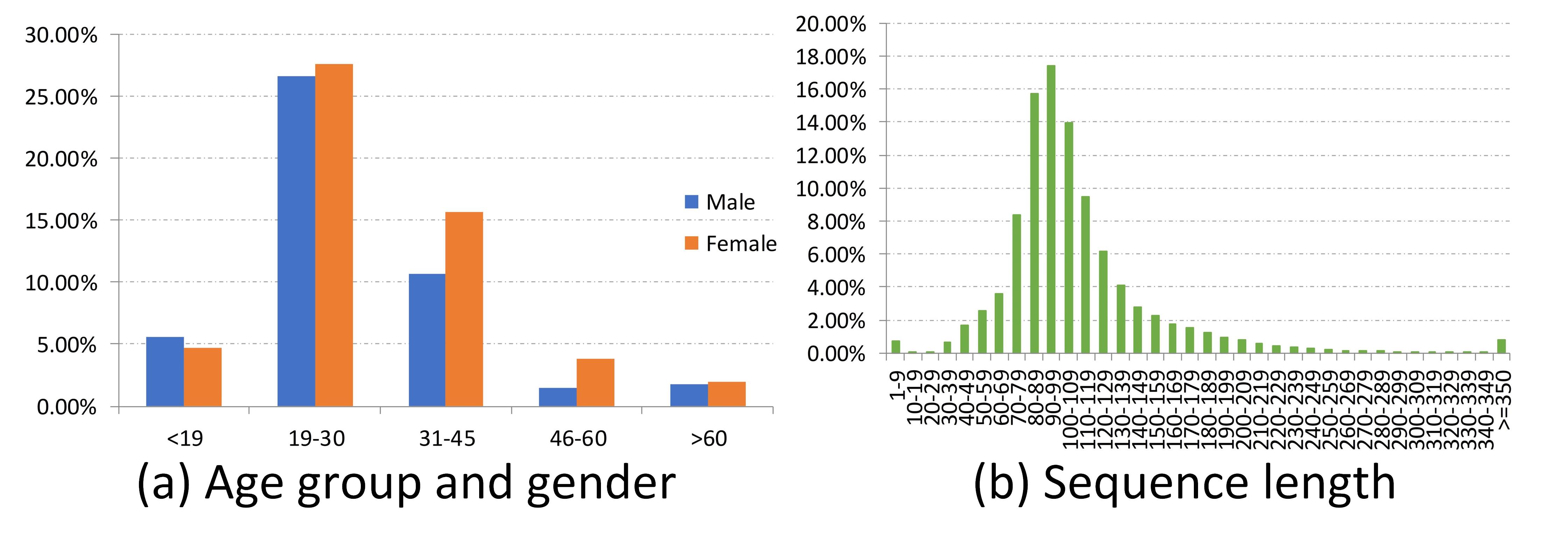}
		\caption{Age group, gender, and  sequence length attributes. Ages are categorized into five groups ($<19$, $19-30$,
			$31-45$, $46-60$, and $>60$). Sequence lengths are grouped at intervals of 10 frames, except for those greater than 350 frames.\footnotesize} 
		\label{figs5}
    \end{figure} 
 
    \subsection{Collection, Statistics and Evaluation}
    \textbf{Collection Process:} To simplify the description, we refer to covariates mentioned in the previous subsection as the ``walking conditions". In the normal walking condition, each subject walks twice. In the remaining 52 walking conditions, each subject only walks once per condition. Therefore, a total of 54 walks per subject are required. Since each subject has to walk 54 times, and the walking conditions have to be changed each time, it takes 2 hours to collect one subject.
	
    \textbf{Dataset statistics:} Figure \ref{figs5} presents the distribution of age, gender, and sequence length in CCGR. The proportions of the various covariates align with the number of walks for each covariate. Furthermore, CCGR exhibits an average of 110 frames per sequence.

    \begin{figure}[t]
		\centering
		\includegraphics[width=1\linewidth]{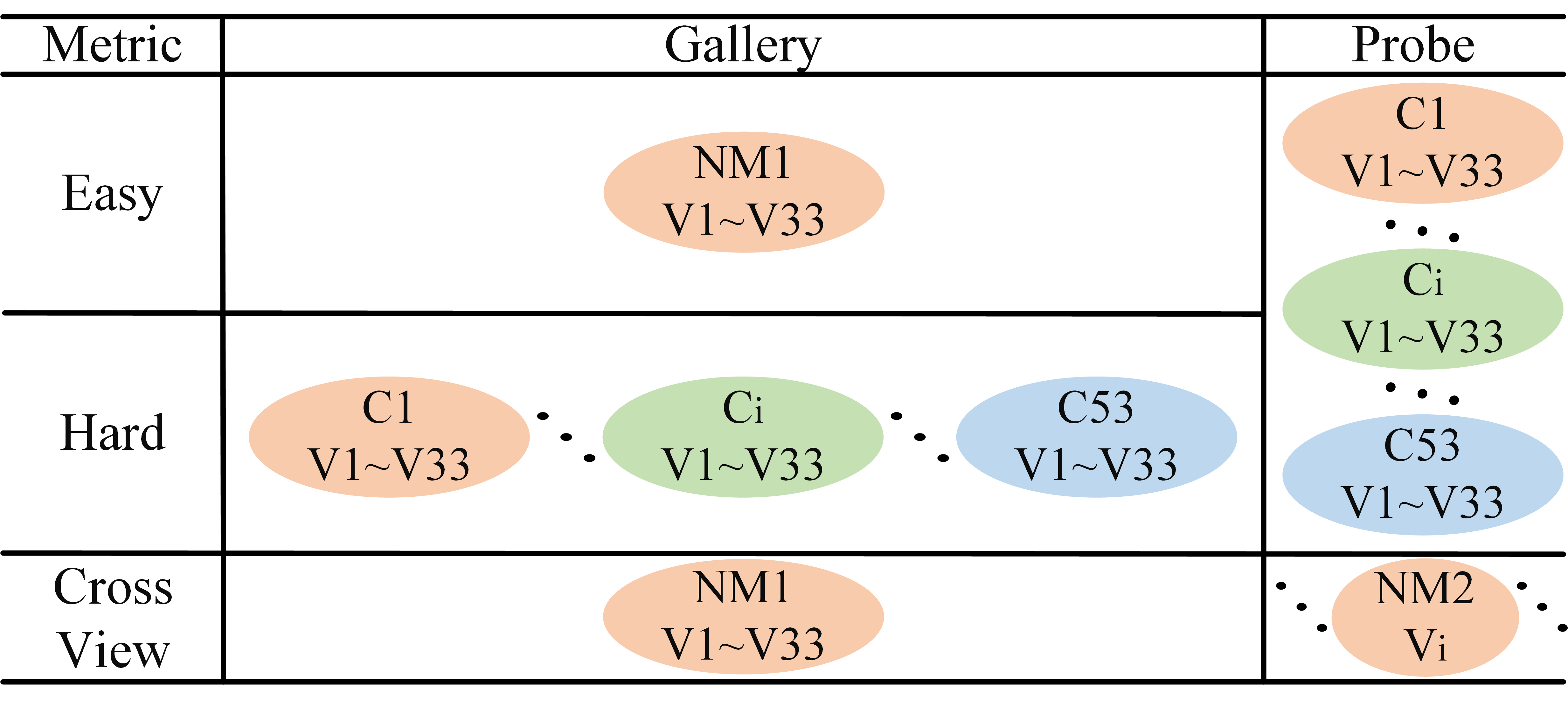}
		\caption{Evaluation metrics. C and V denote covariates and views, where subscripts indicate the order. NM is normal walking. ``Easy" is employed by CASIA-B and OUMVLP (the gallery is normal walking). ``Hard" is similar to GREW and Gait3D, closer to real life (the gallery is uncertain). The cross-view setting inherits from OU-MVLP.\footnotesize} 
		\label{figsx}
    \end{figure} 
    \textbf{Evaluation Protocol:}
    Subjects are labeled from 1 to 1000. Subjects 134 to 164 are missing. Subjects 1 to 600 are used for training, and the rest are used for testing. The evaluation metrics are illustrated in Figure \ref{figsx}. 

    \section{Parsing-based Gait Recognition}
    Although silhouette and pose are commonly employed as gait modalities, they possess significant limitations. Silhouette provides only contour information, while pose offers solely structural details, resulting in sparse and simplistic representations. Consequently, these modalities prove less effective when confronted with complex covariate environments. We are fortunate to discover that parsing can simultaneously provide contour, structural and semantic information. Notably, parsing eliminates texture and color, providing a  basis for treating as a gait pattern.

    Parsing and silhouettes have similar data structures, enabling parsing to inherit all silhouette-based algorithms without modification. This convenient compatibility allows us to explore parsing-based gait recognition efficiently. This paper explores the effectiveness of ``Parsing $+$ silhouette-based algorithms" and calls it \textbf{ParsingGait}.
	
    \section{Baseline on CCGR}
    \subsection{Appearance-based Approaches}
    GEINet \cite{17} adopts a four-layer CNN to learn gait features from GEIs. 
    GaitSet \cite{9} proposes to consider the gait as a set.  
    GaitPart \cite{10} focuses on fine-grained feature extraction and micro-motion feature capture in different body parts. 
    CSTL \cite{66} proposes a temporal modeling module to fuse multi-scale temporal features. 
    GaitGL \cite{68} designs local and global 3D CNNs to extract gait's local and global spatial features. 
    GaitBase \cite{124} provides a structurally simple, empirically powerful, and practically robust baseline model, GaitBase. 
    DeepGaitV2 \cite{126} provides a 22-layer network to address the entire outdoor dataset with many covariates.
	
    \textbf{Implementation details:} All silhouettes are aligned by the approach mentioned in \cite{26} and transformed to 64 $\times$ 44. The batch size is 8 $\times$ 16 $\times$ 30, where 8 denotes the number of subjects, 16 denotes the number of training samples per subject, and 30 is the number of frames. The optimizer is Adam. The number of iterations is 320K. The learning rate starts at 1e-4 and drops to 1e-5 after 200K iterations. For GaitBase and DeepGaitV2: The optimizer is SGD. The number of iterations is 240K. The learning rate starts at 1e-1 and drops by 1$/$10 at 100k, 140k, and 170k. All models are trained on the entire training set; this enables the model to be trained well in some experiments, and in general, with less than 5000 sequences, it is simply impossible to train a larger model well.
    \subsection{Model-based Approaches}
    GaitGraph \cite{32} treats the human skeleton as a graph and then extracts the structural features using a graph convolutional neural network. 
    We train it for 1200 epochs with a batch size of 128. 
    GaitGraph2 \cite{122} proposes a multi-branch graph-based interpretation of gait together with a GCN architecture. 
    We train it for 500 epochs with a batch size of 768. 
    \section{Experiment}
    \begin{table}[t]
	\centering
	\caption{The accuracy of representative methods on CCGR.\footnotesize}
	\resizebox{1.0\columnwidth}{!}{
		\begin{tabular}{c|c|c|c|c|c}
			\toprule[1pt]
			Methods & Year & \textbf{R-1$^{hard}$}  & R-1$^{easy}$       &R-5$^{hard}$    & R-5$^{easy}$ \\ \hline \hline
			GEINet                   & 2016          & \multicolumn{1}{c|}{3.10}  & 4.62            & \multicolumn{1}{c|}{9.20} & 12.7      \\
			GaitSet                  & 2019          & \multicolumn{1}{c|}{25.3}  & 35.3            & \multicolumn{1}{c|}{46.7} & 58.9      \\
			GaitPart                 & 2020          & \multicolumn{1}{c|}{22.6}  & 32.7            & \multicolumn{1}{c|}{42.9} & 55.5     \\
			GaitGL                   & 2021          & \multicolumn{1}{c|}{23.1}  & 35.2            & \multicolumn{1}{c|}{39.9} & 54.1       \\
			CSTL                     & 2021           & \multicolumn{1}{c|}{7.25}  & 11.8            & \multicolumn{1}{c|}{13.79} & 20.1   \\
			GaitBase                 & 2023          & \multicolumn{1}{c|}{31.3}  & 43.8            & \multicolumn{1}{c|}{51.3} & 64.4      \\
			DeepGaitV2               & 2023          & \multicolumn{1}{c|}{42.5} & 55.2           & \multicolumn{1}{c|}{63.2} & 75.2       \\ \hline
			GaitGraph                & 2021                  & \multicolumn{1}{c|}{15.2} & 25.2           & \multicolumn{1}{c|}{37.2} & 51.6       \\
			GaitGraph2               & 2022                & \multicolumn{1}{c|}{0.26} & 0.27           & \multicolumn{1}{c|}{1.4}  & 1.41           \\ \bottomrule[1pt]
		\end{tabular}
	}
	\label{Tab3}
\end{table}

\begin{table}[t]
	\centering
	\caption{The accuracy of ParsingGait on CCGR.\footnotesize}
	
	\resizebox{1.0\columnwidth}{!}{
		\begin{tabular}{c|c|c|c|c|c}
			\hline
			Methods                                                                          & Backbone       & \textbf{R-1$^{hard}$}  & R-1$^{easy}$       & R-5$^{hard}$    & R-5$^{easy}$  \\ \hline \hline
			\multirow{6}{*}{\begin{tabular}[c]{@{}c@{}}\textbf{Parsing}\\ \textbf{Gait}\\ \textbf{(Ours)}\end{tabular}} 
			&GaitSet    & 31.6  & 42.8    & 54.8   & 67         \\ 
			&GaitPart   & 29.0  & 40.9    & 51.5   & 64.5      \\ 
			&GaitGL     & 28.4  & 42.1    & 46.6   & 61.4     \\
			&CSTL       & 27.9  & 40.7    & 47.1   & 61.5      \\ 
			&GaitBase   & 43.2  & 56.9    & 63.7   & 76.0       \\ 
			&DeepGaitV2 & 52.7  & 67.2    & 74.7   & 87.7       \\ \hline
	\end{tabular}}
	\label{Tab4}
\end{table}

\subsection{Analysis of Representative Methods }
    The results are shown in Table \ref{Tab3}. The {R-1$^{hard}$ of GEINet, GaitSet, GaitPart, GaitGL, and CSTL falls below 26\%. While these methods demonstrate near 90\% accuracy on previous indoor datasets, their validity under complex covariates remains untested. In addition, GaitGraph and GaitGraph2 exhibit poorer performance than silhouette-based methods because the pose is sparser than silhouette.

    GaitBase and DeepGaitV2 are proposed to address the challenge of outdoor datasets; they are more robust against complex covariates. However, \textbf{DeepGaitV2 achieves an impressive 82\% rank-1 accuracy on the outdoor dataset GREW. In contrast, its performance on CCGR falls considerably below, reaching a mere 43\%}. This disparity may be due to the lack of individual-level diversity in the existing outdoor datasets.
            \begin{figure}[t]
    	\centering 
    	\setlength{\belowcaptionskip}{-0.3cm} 
    	\includegraphics[width=1\linewidth]{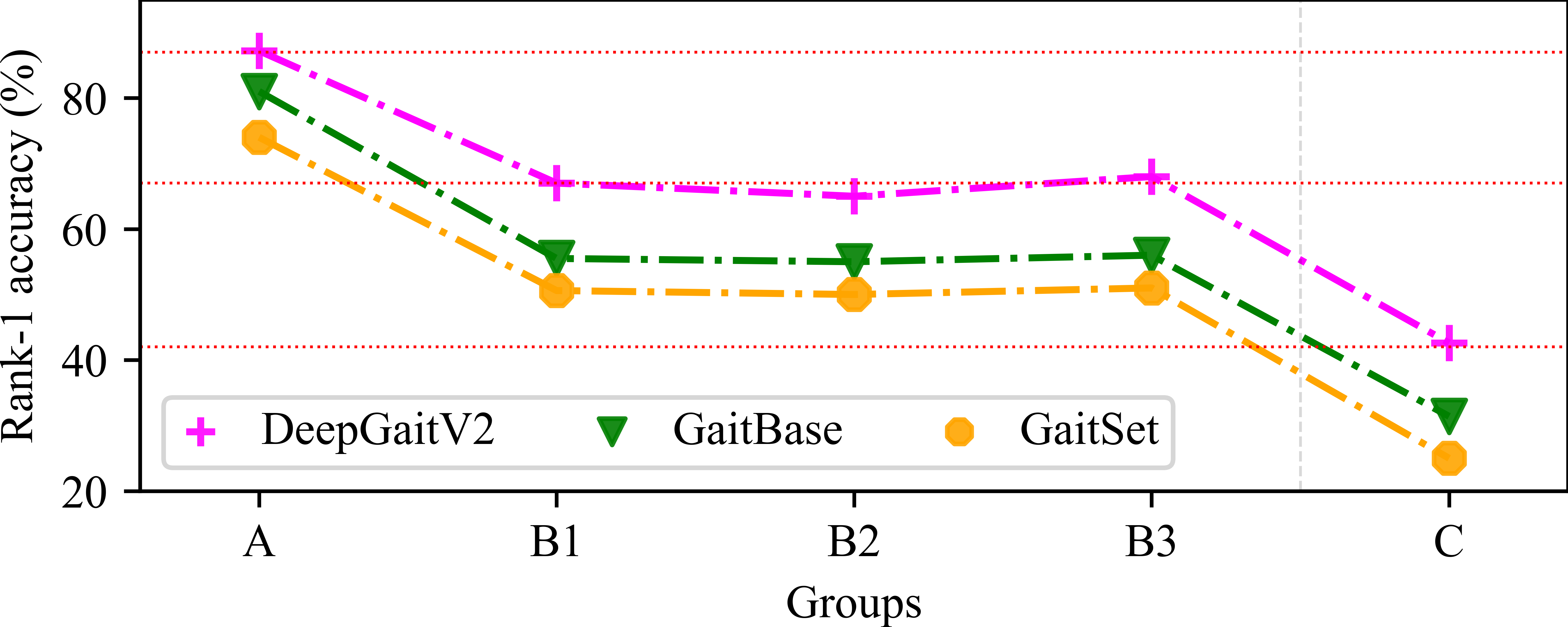}
    	\caption{Impact of population and individual diversity.\footnotesize}
    	\label{figs6}
    \end{figure}
        \begin{table}[t]
		\centering
		\caption{\textbf{Covariate Sampling Setup}. Seq, Sbj, and NoC refer to the sequence, subject, and number of covariates.
  \footnotesize}
		\resizebox{1.0\columnwidth}{!}{
            \begin{tabular}{c|c|c|c|c}
           \toprule[1pt]
               Groups & Similar to & Sample Setup  & \begin{tabular}[c]{@{}c@{}}NoC per \\ Sbj\end{tabular} & \begin{tabular}[c]{@{}c@{}}NoC of the\\ Sub-dataset\end{tabular} \\ \hline
            A    & CASIA-B  & NM, BG, CL, Layer1            & 3                                                          & 3                                                             \\ \hline
            B1   &\multirow{3}{*}{\begin{tabular}[c]{@{}c@{}}Gait3D/\\ GREW\end{tabular}}   &  random 8 Seqs per Sbj & Max 8                                                      & 53                                                            \\ 
            B2   &    &  random 8 Seqs per Sbj & Max 8                                                      & 53                                                           \\ 
            B3   &      &  random 8 Seqs per Sbj & Max 8                                                      & 53                                                            \\ \hline
          C   & Ours   & Use all sequences                       & 53                                                        & 53                                                            \\ \bottomrule[1pt]
        \end{tabular}
	       }
	\label{Tab5}
    \end{table}

    \subsection{Analysis of Parsing-based Gait Recognition}
    As shown in Table \ref{Tab4}, the accuracy of ParsingGait is substantially improved. These findings effectively illustrate the three main advantages of parsing: feasibility, validity, and compatibility. By distinguishing between different body parts, parsing makes it more robust in the face of complex covariates. ParsingGait is the same computationally efficient as its silhouette-based counterpart because our parsing is consistent with the silhouette data structure.

    \subsection{Population and Individual-Level Diversity}
    We research the impact of covariate diversity by sampling and isolating various covariates. The specific sampling setup is provided in Table \ref{Tab5}. The experiments are categorized into five groups. Group A represents the absence of covariate diversity, while Group B demonstrates population-level diversity without individual-level diversity. Lastly, Group C exhibits both population-level and individual-level diversity. To simulate the short-term collectio, Group B is sampled in 2 specific ways: 1) One covariate per ID is randomly sampled, and then eight views under that covariate were randomly sampled. 2) One view per ID is randomly sampled, and then eight covariates under that view are randomly sampled. During the operation, two ways are randomly selected.

    Based on the experimental data in Figure \ref{figs6}. From A to B1/2/3, the accuracy averagely decreased by -18.6\%. However, from B1/2/3 to C, the accuracy averagely decreased by -25.1\%. These findings indicate that \textbf{relying solely on population-level diversity is insufficient to accurately represent the underlying challenge, while individual-level diversity also is a significant challenge}. In addition, the trend of Figure \ref{figs6} is generally consistent with Figure \ref{figs1} at the beginning of the paper, further strengthening the credibility of the experimental results.  

    \begin{figure}[t]
    	\centering
    	\includegraphics[width=1\linewidth]{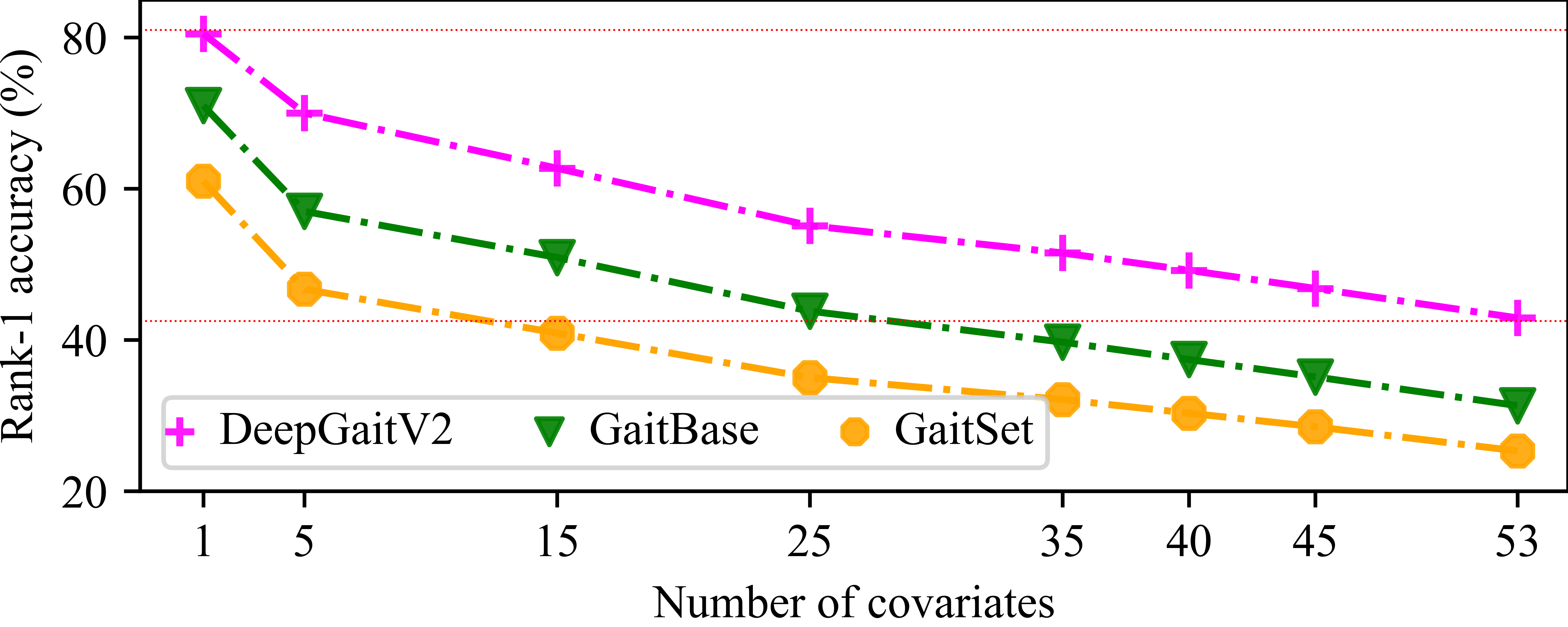}
    	\caption{Impact of the number of covariates.\footnotesize}
    	\label{figs7}
    \end{figure}

    \subsection{Impact of the Number of Covariates}
    We examine how the number of covariates impacts accuracy, and the experimental outcomes are illustrated in Figure \ref{figs7}. The accuracy is substantially decreased as we progressively increase the covariate number from 1 to 53. Furthermore, a troubling trend emerges: even when the number reaches 53, the decline in accuracy rate does not significantly decelerate. This observation may indicate that gait recognition faces greater challenges in real-world scenarios, hinting at potential obstacles in open-world conditions.

\begin{table}[ht!]
          \centering
          \setlength{\abovecaptionskip}{0.2cm}
          \caption{\textbf{Single-Covariate Evaluation}: R-1$^{easy}$ accuracy (\%) with excluding identical-view cases. $\downarrow$ and \textbf{bold} respectively indicate the sub-\textit{Average} and SOTA performance.\footnotesize}
          \resizebox{1\columnwidth}{!}{
    \begin{tabular}{c|c|c|c|c|c}
    \toprule
\multicolumn{6}{c}{Gallery: Normal 1}                                                                                                 \\ \midrule
\multicolumn{3}{c|}{Publication}                                                                 & CVPR23 & Arxiv23  & Ours        \\ \midrule
Type                                                                 & Covariate    & Abbr. & GaitBase & DeepGaitV2 & ParsingGait \\ \hline \midrule
\multirow{8}{*}{Carrying}                                                & Book         & BK    & 65.7     & 75.3       & 85.5        \\
                                                                         & Bag          & BG    & 64.9     & 75.4       & 86.1        \\
                                                                         & Heavy Bag    & HVBG  & 60.0$\downarrow$     & 72.3       & 84.2        \\
                                                                         & Box          & BX    & 61.5     & 71.6       & 83.0        \\
                                                                         & Heavy Box    & HVBX  & 58.7$\downarrow$     & 69.7$\downarrow$       & 81.9$\downarrow$        \\
                                                                         & Trolley Case & TC    & 64.1     & 73.0       & 83.4        \\
                                                                         & Umbrella     & UB    & 47.2$\downarrow$     & 60.5$\downarrow$       & 71.3$\downarrow$        \\ \cmidrule{2-6}
                                                                         & \textit{Average}      & -     & 60.3     & 71.1       & \textbf{82.2}        \\ \hline \midrule
Clothing                                                                 & Thick Coat   & CL    & 40.4     & 53.5       & 66.8        \\ \hline \midrule
\multirow{8}{*}{Road}                                                    & Up Ramp      & UTR   & 60.3$\downarrow$     & 69.5$\downarrow$       & 80.9        \\
                                                                         & Down Ramp    & DTR   & 60.5$\downarrow$     & 70.1$\downarrow$       & 80.2        \\
                                                                         & Up Stair     & UTS   & 54.9$\downarrow$     & 66.7$\downarrow$       & 78.0$\downarrow$        \\
                                                                         & Down Stair   & DTS   & 54.0$\downarrow$     & 65.4$\downarrow$       & 76.7$\downarrow$        \\
                                                                         & Bumpy Road   & BM    & 63.3     & 71.4       & 82.0        \\
                                                                         & Curved Road  & CV    & 70.0     & 77.3       & 86.1        \\
                                                                         & Soft Road    & SF    & 66.0     & 73.2       & 83.7        \\ \cmidrule{2-6}
                                                                         & \textit{Average}      & -     & 61.3     & 70.5       & \textbf{79.3}        \\ \hline \midrule
\multirow{4}{*}{Speed}                                                   & Normal 1     & NM1   & 76.6     & 83.5       & 91.3        \\
                                                                         & Fast         & FA    & 47.2$\downarrow$     & 60.7$\downarrow$       & 74.1$\downarrow$        \\
                                                                         & Stationary   & ST    & 32.0$\downarrow$     & 45.0$\downarrow$       & 60.9$\downarrow$        \\ \cmidrule{2-6}
                                                                         & \textit{Average}      & -     & 51.9     & 63.1       & \textbf{75.4}        \\ \hline \midrule
\multirow{5}{*}{\begin{tabular}[c]{@{}c@{}}Walking\\ Style\end{tabular}} & Normal 2     & NM2   & 75.3     & 82.3       & 90.7        \\
                                                                         & Confident    & CF    & 64.9     & 74.8       & 83.9        \\
                                                                         & Freedom       & FD    & 57.1     & 68.1       & 79.2        \\
                                                                         & Multi-person        & MP    & 24.0$\downarrow$     & 32.6$\downarrow$       & 39.4$\downarrow$        \\ \cmidrule{2-6}
                                                                         & \textit{Average}      & -     & 55.3     & 64.4       & \textbf{73.3}        \\ \bottomrule
\end{tabular}
    }
    \label{Tab-3}
    \end{table}
\begin{table}[ht!]
          \centering
          \caption{\textbf{Mixed-Covariate Evaluation}: R-1$^{easy}$ accuracy (\%) with excluding identical-view cases. 
          We use ``-" to connect the mixed covariates. Tab.~\ref{Tab-3} presents the dictionary containing abbreviations and their corresponding full spellings of these covariates. 
          $\downarrow$ and \textbf{bold} respectively indicate the sub-\textit{Average} and SOTA performance. 
          \footnotesize}
          \resizebox{1\columnwidth}{!}{
    \begin{tabular}{c|c|c|c|c}
    \toprule
\multicolumn{5}{c}{Gallery: Normal 1}                                                                                              \\ \midrule
\multicolumn{2}{c|}{Publication}                                                          & CVPR23 & Arxiv23  & Ours          \\ \midrule
Category                                                                & Covariate       & GaitBase & DeepGaitV2 & ParsingGait   \\ \hline \midrule
\multirow{12}{*}{\begin{tabular}[c]{@{}c@{}}Two\\ Mixed\end{tabular}}   & CL-UB           & 25.2$\downarrow$     & 37.8$\downarrow$       & 46.9$\downarrow$          \\
                                                                        & HVBX-BG         & 52.1     & 64.7       & 78.3          \\
                                                                        & BG-TC           & 58.1     & 69.3       & 81.3          \\
                                                                        & SF-CL           & 36.1$\downarrow$     & 48.0$\downarrow$       & 62.8$\downarrow$          \\
                                                                        & UTR-BX          & 51.0     & 62.0       & 75.4          \\
                                                                        & DTR-BK          & 55.1     & 66.0       & 77.4          \\
                                                                        & DTS-HVBX        & 42.6$\downarrow$     & 56.1$\downarrow$       & 69.8$\downarrow$          \\
                                                                        & UTS-BG          & 46.8     & 60.9       & 74.5          \\
                                                                        & BM-CL           & 35.2$\downarrow$     & 46.3       & 61.8          \\
                                                                        & CV-HVBX         & 61.0     & 70.8       & 82.0          \\
                                                                        & CL-CF           & 39.2$\downarrow$     & 52.7$\downarrow$       & 65.6$\downarrow$          \\ \cmidrule{2-5}
                                                                        & \textit{Average}         & 45.7     & 57.7       & \textbf{70.5}          \\ \hline \midrule
\multirow{12}{*}{\begin{tabular}[c]{@{}c@{}}Three\\ Mixed\end{tabular}} & CL-UB-BG        & 23.4$\downarrow$     & 36.1$\downarrow$       & 44.9$\downarrow$          \\
                                                                        & BX-BG-CL        & 35.1$\downarrow$     & 48.8       & 60.7          \\
                                                                        & BG-TC-CL        & 34.3$\downarrow$     & 48.5       & 63.0          \\
                                                                        & SF-UB-BG        & 36.4$\downarrow$     & 49.4       & 62.5          \\
                                                                        & UTR-HVBX-CL     & 31.8$\downarrow$     & 43.1$\downarrow$       & 55.3$\downarrow$          \\
                                                                        & DTR-BK-BG       & 49.2     & 61.7       & 74.9          \\
                                                                        & DTS-HVBX-CL     & 26.4$\downarrow$     & 38.0$\downarrow$       & 49.1$\downarrow$          \\
                                                                        & UTS-BG-CL       & 25.1$\downarrow$     & 37.7$\downarrow$       & 52.5$\downarrow$          \\
                                                                        & BM-CL-BG        & 33.0$\downarrow$     & 44.8$\downarrow$       & 59.6$\downarrow$          \\
                                                                        & CV-BX-BG        & 58.8     & 69.6       & 80.8          \\
                                                                        & UB-BG-FA        & 28.0$\downarrow$     & 41.0$\downarrow$       & 52.8$\downarrow$          \\ \cmidrule{2-5}
                                                                        & \textit{Average}         & 34.7     & 47.1       & \textbf{59.7}          \\ \hline \midrule
\multirow{9}{*}{\begin{tabular}[c]{@{}c@{}}Four\\ Mixed\end{tabular}}   & CL-UB-BG-FA     & 16.2$\downarrow$     & 27.6$\downarrow$       & 35.7$\downarrow$          \\
                                                                        & BM-CL-BG-BX     & 32.2     & 43.5       & 56.1          \\
                                                                        & BG-TC-CL-CV     & 38.0     & 51.2       & 66.9          \\
                                                                        & DTR-BK-BG-CL    & 32.2     & 44.9       & 56.9          \\
                                                                        & DTS-BX-CL-BG    & 25.6     & 37.3       & 48.9          \\
                                                                        & SF-UB-BG-CL     & 20.6$\downarrow$     & 31.8$\downarrow$       & 41.9$\downarrow$          \\
                                                                        & BG-TC-CL-ST     & 11.7$\downarrow$     & 18.4$\downarrow$       & 29.4$\downarrow$          \\
                                                                        & UTS-UB-BG-CL    & 15.8$\downarrow$     & 26.1$\downarrow$       & 36.4$\downarrow$          \\ \cmidrule{2-5}
                                                                        & \textit{Average}         & 24.0     & 35.1       & \textbf{46.5}          \\ \hline \midrule
\multirow{4}{*}{\begin{tabular}[c]{@{}c@{}}Five\\ Mixed\end{tabular}}   & \multirow{2}{*}{\begin{tabular}[c]{@{}c@{}}BG-TC-CL-\\ CV-UB\end{tabular}}  & \multirow{2}{*}{34.1}     & \multirow{2}{*}{35.9}       & \multirow{2}{*}{\textbf{47.4}}          \\ 
& & & & \\ \cmidrule{2-5}
                                                                        & \multirow{2}{*}{\begin{tabular}[c]{@{}c@{}}UTR-BG-CL-\\ BX-CV\end{tabular}} & \multirow{2}{*}{31.3}     & \multirow{2}{*}{45.2}       & \multirow{2}{*}{\textbf{58.3}}          \\ 
& & & & \\ \bottomrule
\end{tabular}
    }
    \label{Tab-1}
    \end{table}
\begin{table}[ht!]
         \centering
         \caption{\textbf{Cross-View Evaluation}: Rank-1 accuracy (\%) with excluding identical-view cases. \footnotesize}
         \resizebox{1.0\columnwidth}{!}{
        \begin{tabular}{c|c|c|c|c}
        \toprule
\multicolumn{5}{c}{Cross-view Evaluation}                               \\ \midrule
\multicolumn{2}{c|}{Publications}       & CVPR23 & Arxiv23  & Ours        \\ \midrule
\multicolumn{1}{c|}{\multirow{2}{*}{\begin{tabular}[c]{@{}c@{}}Pitch \\ Angle \\ \end{tabular}}} & \multicolumn{1}{c|}{\multirow{2}{*}{\begin{tabular}[c]{@{}c@{}}Probe \\ View \\ \end{tabular}}}  & \multirow{2}{*}{GaitBase} & \multirow{2}{*}{DeepGaitV2} & \multirow{2}{*}{ParsingGait} \\ 
 & & & & \\ \hline \midrule
\multirow{10}{*}{5$^\circ$} & 0.0$^\circ$        & 80.1     & 85.7       & 90.6        \\
                     & 22.5$^\circ$       & 84.7     & 89.5       & 93.1        \\
                     & 45.0$^\circ$       & 83.7     & 89.1       & 93.9        \\
                     & 67.5$^\circ$       & 79.3     & 85.7       & 93.6        \\
                     & 90.0$^\circ$       & 75.7     & 83.7       & 93.2        \\
                     & 112.5$^\circ$      & 76.9     & 84.6       & 93.2        \\
                     & 135.0$^\circ$      & 81.6     & 87.1       & 93.7        \\
                     & 157.5$^\circ$      & 83.8     & 88.6       & 92.7        \\
                     & 180.0$^\circ$      & 77.4     & 83.3       & 89.9        \\ \cmidrule{2-5}
                     & \textit{Average}    & 80.4     & 86.4       & 92.6        \\ \hline \midrule 
\multirow{10}{*}{30$^\circ$} & 0.0$^\circ$        & 79.6     & 85.2       & 92.0        \\
                     & 22.5$^\circ$       & 85.0     & 89.8       & 93.6        \\
                     & 45.0$^\circ$       & 86.0     & 90.9       & 94.9        \\
                     & 67.5$^\circ$       & 82.7     & 88.8       & 95.0        \\
                     & 90.0$^\circ$       & 78.9     & 86.4       & 94.6        \\
                     & 112.5$^\circ$      & 79.1     & 86.3       & 94.5        \\
                     & 135.0$^\circ$      & 82.8     & 88.5       & 94.5        \\
                     & 157.5$^\circ$      & 84.1     & 89.9       & 93.7        \\
                     & 180.0$^\circ$      & 79.5     & 85.3       & 91.8        \\ \cmidrule{2-5}
                     & \textit{Average}    & 82.0     & 87.9       & 93.9        \\ \hline \midrule 
\multirow{10}{*}{55$^\circ$} & 0.0$^\circ$        & 74.8     & 81.8       & 90.6        \\ 
                     & 22.5$^\circ$       & 81.5     & 86.7       & 93.3        \\
                     & 45.0$^\circ$       & 83.9     & 88.9       & 95.0        \\
                     & 67.5$^\circ$       & 82.2     & 88.4       & 95.1        \\
                     & 90.0$^\circ$       & 63.6     & 76.3       & 92.0        \\
                     & 112.5$^\circ$      & 77.3     & 84.5       & 93.6        \\
                     & 135.0$^\circ$      & 81.2     & 87.4       & 93.9        \\
                     & 157.5$^\circ$      & 80.8     & 86.3       & 93.2        \\
                     & 180.0$^\circ$      & 75.9     & 83.2       & 91.3        \\ \cmidrule{2-5}
                     & \textit{Average}    & 77.9     & 84.8       & 93.1        \\ \hline \midrule 
\multirow{6}{*}{75$^\circ$}  & 0.0$^\circ$        & 64.4     & 74.8       & 86.0        \\
                     & 45.0$^\circ$       & 78.7     & 85.2       & 92.7        \\
                     & 90.0$^\circ$       & 40.8     & 60.9       & 87.5        \\
                     & 135.0$^\circ$      & 73.2     & 80.5       & 90.6        \\
                     & 180.0$^\circ$      & 62.5     & 74.0       & 86.2        \\ \cmidrule{2-5}
                     & \textit{Average}    & 63.9     & 75.1       & 88.6        \\ \hline \midrule 
OverHead             & -          & 2.0      & 8.4        & 32.0        \\ \bottomrule
\end{tabular}
         \label{tab-2}
          }
    \end{table}

    \subsection{Evaluation of Covariates and Views}
    In Table \ref{Tab3} and \ref{Tab4}, we evaluate the overall performance of CCGR.  Next, we analyze the impact of different covariates and views using the ``easy" evaluation criteria. Covariates in Table  \ref{Tab-3} and \ref{Tab-1} that are challenging and prone to causing failure have been denoted with $\downarrow$.
    
    \textbf{Single-Covariate Evaluation:} As shown in Table \ref{Tab-3}. Multi-person walking significantly affects accuracy because many parts of the human body are obscured. Speed also significantly affects the accuracy as it dramatically impacts the temporal feature extraction of the algorithm. Clothing is still a big challenge. In addition, carrying and road also have a notable negative impact on accuracy.

    \textbf{Mixed-Covariate Evaluation:} As shown in Table \ref{Tab-1}. Mixed covariates impact precision more, with a significant classical decrease as the number of mixes increases. for example, ``Bag $\rightarrow$ BG-TC $\rightarrow$ BG-TC-CL $\rightarrow$ BG-TC-CL-ST", accuracy is gradually declining. However, mixed covariates are a challenge that must be addressed because ideal conditions for single covariates in real life tend to be rare.
    
    \textbf{Cross-View Evaluation:} As shown in Table \ref{tab-2}. The existing algorithms perform well, considering only the views. The current challenge with views is how to address the high-pitch angle case. Encouragingly, ParsingGait demonstrates distinct improvement in recognizing overhead views, indicating that addressing high-pitch angles is promising.
    
    \section{Conclusion}
    
     This paper introduces CCGR, a well-labeled dataset consisting of over one million sequences, which provides diversity at both the population and individual levels. Our experiments demonstrate that individual-level diversity is as challenging as population-level diversity.  As gait recognition on many public gait datasets is close to saturation, our dataset CCGR introduces more challenges, specifically more covariates, into gait recognition. Future works can explore how gait recognition is affected by different covariates and how to design robust gait recognition.  

\section{Acknowledgements}
This work wad supported by the Natural Science Foundation of Hunan Province (No.2023JJ30697), the Changsha Natural Science Foundation (No.kq2208286) and the National Natural Science Foundation of China (No.61502537). This work was also supported in part by the National Key Research, and in part by Development Program of China under Grant (No.61976144) and the Shenzhen International Research Cooperation Project under Grant (No.GJHZ20220913142611021).
\section{Additional}

\subsection{Ethical Statement}
All subjects are openly recruited and participate voluntarily, and they need to read and acknowledge the collection protocol by signature and fingerprint. In return, we pay each subject a fee for data copyright.

\subsection{Analysis of Silhouette Extraction}
    Earlier work \cite{25} uses \textbf{background subtraction} \cite{115} to obtain silhouettes, but background subtraction requires heuristic pre-processing and is unsuitable for large-scale practical use. Recently, \textbf{GREW} \cite{76} uses the instance segmentation (\textbf{Ins-Seg}) algorithm HTC \cite{109} to obtain silhouettes, and \textbf{Gait3D} \cite{75} uses the semantic segmentation (\textbf{Sem-Seg}) algorithm HRNet-segmentation \cite{112} to obtain silhouettes. There are also unused panoramic segmentation (\textbf{Pan-Seg}) methods. We compare instance, semantic, and panoramic segmentation, except for background subtraction, which requires heuristic processing. For the three types of methods, we choose 3 SOTA algorithms: HTC \cite{109}, Segformer \cite{110}, and PanopticFPN \cite{114}.
    
    Next, works \cite{120,121} use the obtained parsing maps to improve the original silhouette, but their improved approach has heuristic processing, which may be less robust in complex environments. Work \cite{120} uses manual parsing and is unusable on a large scale. We compare the improved silhouette (\textbf{Imp-s}) of work \cite{121}. The above works do not propose using the parsing directly as gait data; they only improve the original silhouette.
    
    To save time, we conduct experiments on the sub-CCGR (Subjects 1 to 60 are used for training, and subjects 961 to 100 are used for testing); this is sufficient for the comparison. The results are listed in figure \ref{Fig1}.
    \begin{figure}[t]
   	\centering 	
    	\includegraphics[width=1\linewidth]{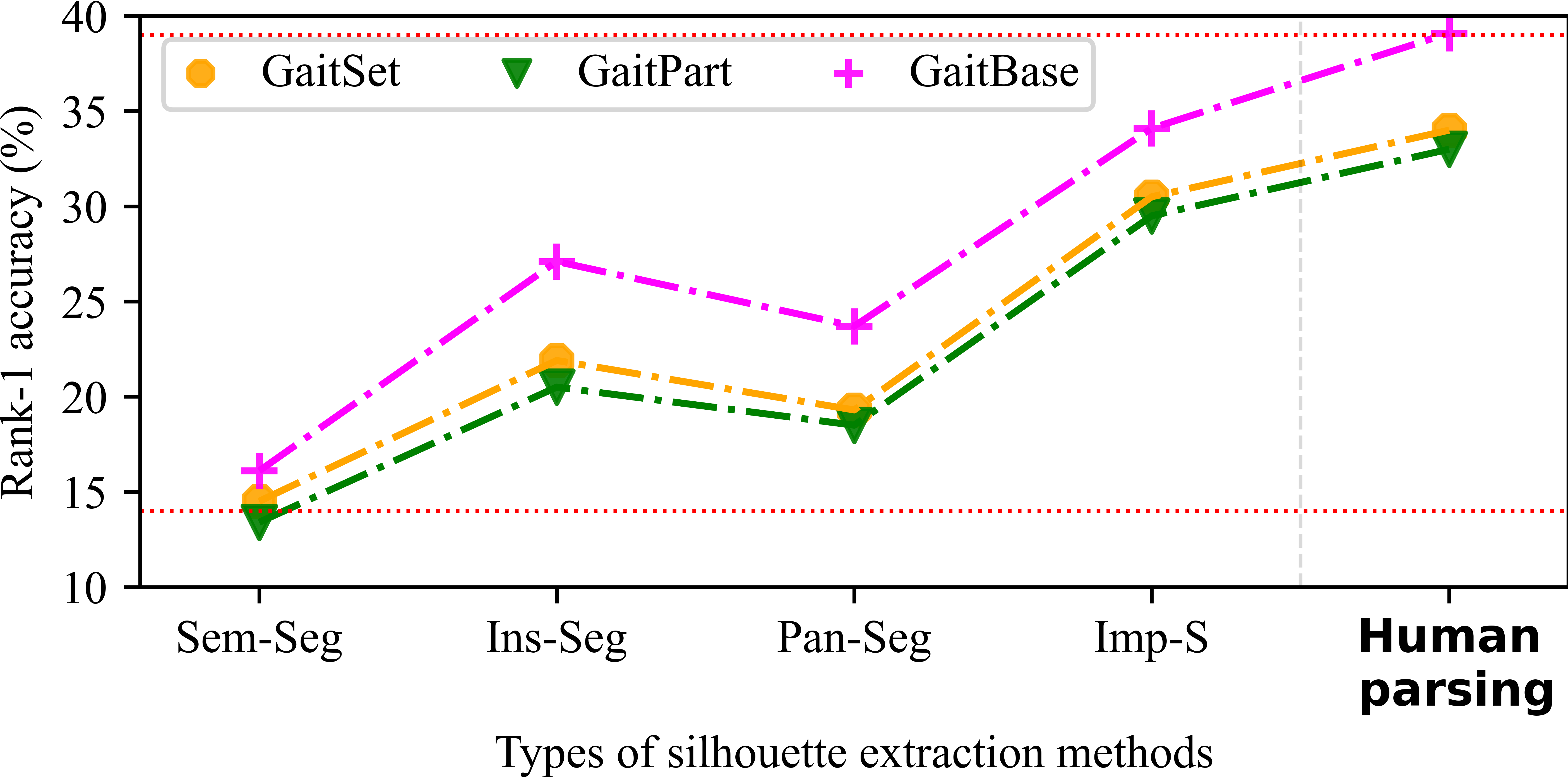}
    	\caption{Rank-1 accuracy (\%) on the sub-CCGR when using different types of silhouette extraction methods. Sem-Seg, Ins-Seg, Pan-Seg, and Imp-s mean semantic segmentation, instance segmentation, panoramic segmentation, and improved silhouette.\footnotesize}
  	\label{Fig1}
    \end{figure}
    \begin{figure}[h]
	\centering	
	\includegraphics[width=1\linewidth]{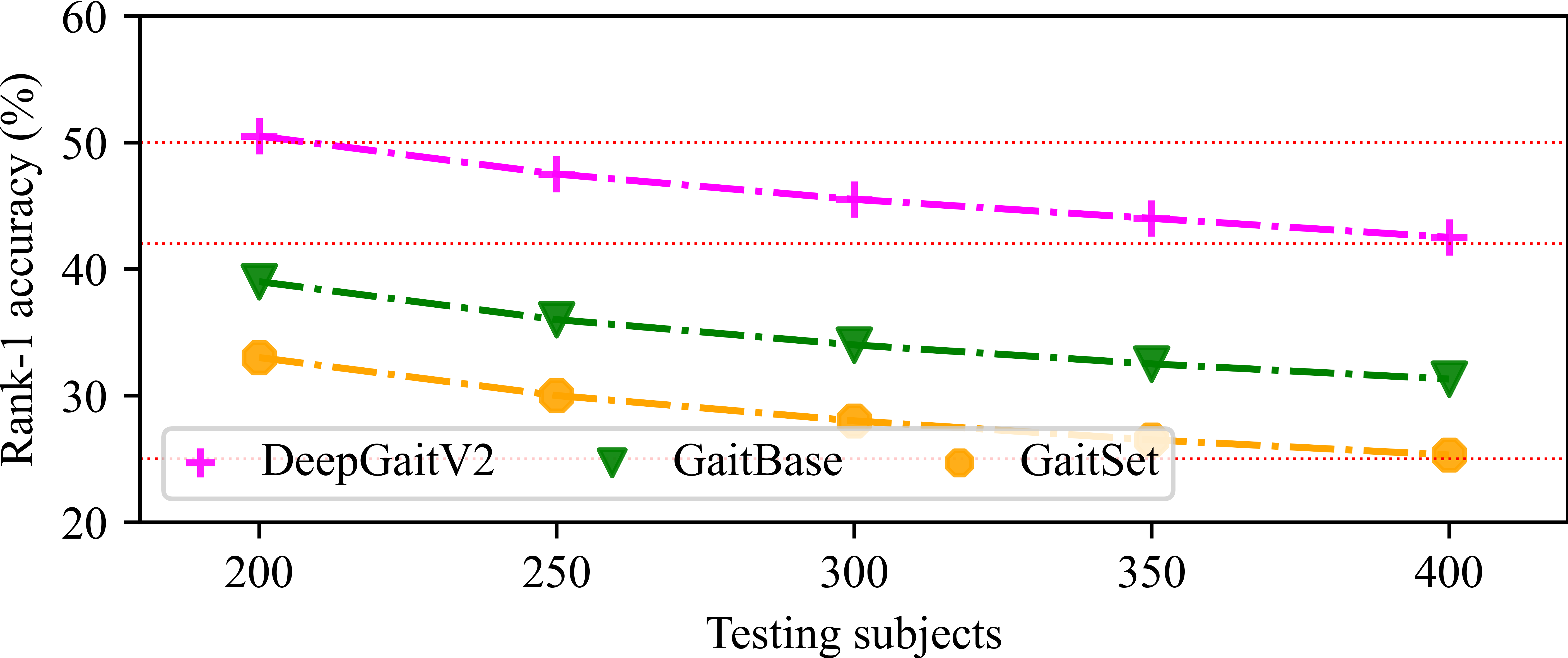}
	\caption{Rank-1 accuracy (\%) with different numbers of testing subjects.\footnotesize}
	\label{Fig2}
    \end{figure}
    
    There is a significant improvement in accuracy when using human parsing. However, recognition accuracy is lower when using HTC, Segformer, and PanopticFPN because they are not initially designed for human segmentation. Improved silhouette (Imp-S) performs poorly, suggesting that this method is not robust to covariates.
    
    Using more accurate silhouettes for CCGR remains very challenging, illustrating that \textbf{the diversity of covariates at the population and individual level} is the main reason for the challenge of the CCGR.

\subsection{Impact of the number of testing subjects}
    Figure \ref{Fig2} shows the accuracy as the testing subjects decrease from 400 to 200. There is a slight improvement in accuracy as the number of testing subjects decreases. However, even if the testing subjects are reduced to 200, the accuracy is still relatively low, illustrating that \textbf{the diversity of covariates at the population and individual level} is the main reason for the challenge of the CCGR.

\subsection{Evaluation of Covariates with ``Hard" Metric}
In the main text, we evaluate the covariates using the ``easy" metric, and here we give a further evaluation under the "hard" metric. The results for "Single-Covariate Evaluation" and "Mixed-Covariate Evaluation" are shown in Table \ref{Tab-3S} and Table \ref{Tab-4S}, respectively.

Hard metrics are more difficult than easy metrics, so there is a decrease in accuracy, but the trend in the impact of covariates does not change. Multi-person walking, speed, clothing, carrying, and road significantly affect gait. In the speed type, stationary had the most significant impact; in the carrying type, umbrellas had the most significant impact; and in the road type, stairs had the most significant impact.

Mixed covariates are more complex than single covariates because they introduce multiple changes simultaneously. With two-mixed, three-mixed, four-mixed, and five-mixed, the precision has gradually decreased. However, this is one of the challenges that must be addressed because natural walks often have multiple covariates simultaneously.

\begin{table}[t!]
          \centering
          \setlength{\abovecaptionskip}{0.2cm}
          \caption{\textbf{Single-Covariate Evaluation}: R-1$^{hard}$ accuracy (\%) with excluding identical-view cases. $\downarrow$ and \textbf{bold} respectively indicate the sub-\textit{Average} and state-of-the-art performance.\footnotesize}
          \resizebox{1\columnwidth}{!}{
\begin{tabular}{cccccc}
\toprule
\multicolumn{6}{c}{Hard Metric}                                                                                                                                                                                                                        \\ \midrule
\multicolumn{3}{c|}{Publication}                                                                                                                                & \multicolumn{1}{c|}{CVPR23}   & \multicolumn{1}{c|}{Arxiv23}    & Ours        \\ \midrule
\multicolumn{1}{c|}{Type}                                                                      & \multicolumn{1}{c|}{Covariate}    & \multicolumn{1}{c|}{Abbr.} & \multicolumn{1}{c|}{GaitBase} & \multicolumn{1}{c|}{DeepGaitV2} & ParsingGait \\ \hline \midrule
\multicolumn{1}{c|}{\multirow{8}{*}{Carrying}}                                                 & \multicolumn{1}{c|}{Book}         & \multicolumn{1}{c|}{BK}    & \multicolumn{1}{c|}{42.0}     & \multicolumn{1}{c|}{53.6}       & 65.9        \\
\multicolumn{1}{c|}{}                                                                          & \multicolumn{1}{c|}{Bag}          & \multicolumn{1}{c|}{BG}    & \multicolumn{1}{c|}{41.0}     & \multicolumn{1}{c|}{52.7}       & 65.3        \\
\multicolumn{1}{c|}{}                                                                          & \multicolumn{1}{c|}{Heavy Bag}    & \multicolumn{1}{c|}{HVBG}  & \multicolumn{1}{c|}{38.4$\downarrow$ }    & \multicolumn{1}{c|}{50.1$\downarrow$}       & 63.6        \\
\multicolumn{1}{c|}{}                                                                          & \multicolumn{1}{c|}{Box}          & \multicolumn{1}{c|}{BX}    & \multicolumn{1}{c|}{40.9}     & \multicolumn{1}{c|}{51.9}       & 64.3        \\
\multicolumn{1}{c|}{}                                                                          & \multicolumn{1}{c|}{Heavy Box}    & \multicolumn{1}{c|}{HVBX}  & \multicolumn{1}{c|}{39.7}     & \multicolumn{1}{c|}{51.1}       & 63.5        \\
\multicolumn{1}{c|}{}                                                                          & \multicolumn{1}{c|}{Trolley Case} & \multicolumn{1}{c|}{TC}    & \multicolumn{1}{c|}{41.3}     & \multicolumn{1}{c|}{52.7}       & 65.4        \\
\multicolumn{1}{c|}{}                                                                          & \multicolumn{1}{c|}{Umbrella}     & \multicolumn{1}{c|}{UB}    & \multicolumn{1}{c|}{27.8$\downarrow$}     & \multicolumn{1}{c|}{40.0$\downarrow$}      & 51.0$\downarrow$       \\ \cmidrule{2-6}
\multicolumn{1}{c|}{}                                                                          & \multicolumn{1}{c|}{Average}      & \multicolumn{1}{c|}{-}     & \multicolumn{1}{c|}{38.7}     & \multicolumn{1}{c|}{50.3}       & \textbf{62.7}        \\ \hline \midrule

\multicolumn{1}{c|}{Clothing}                                                                  & \multicolumn{1}{c|}{Thick Coat}   & \multicolumn{1}{c|}{CL}    & \multicolumn{1}{c|}{32.1}     & \multicolumn{1}{c|}{44.3}       & \textbf{56.5}        \\ \hline \midrule

\multicolumn{1}{c|}{\multirow{8}{*}{Road}}                                                     & \multicolumn{1}{c|}{Up Ramp}      & \multicolumn{1}{c|}{UTR}   & \multicolumn{1}{c|}{39.7}     & \multicolumn{1}{c|}{50.7}       & 63.2        \\
\multicolumn{1}{c|}{}                                                                          & \multicolumn{1}{c|}{Down Ramp}    & \multicolumn{1}{c|}{DTR}   & \multicolumn{1}{c|}{39.2$\downarrow$ }     & \multicolumn{1}{c|}{50.6$\downarrow$ }       & 61.9$\downarrow$         \\
\multicolumn{1}{c|}{}                                                                          & \multicolumn{1}{c|}{Up Stair}     & \multicolumn{1}{c|}{UTS}   & \multicolumn{1}{c|}{36.9$\downarrow$ }     & \multicolumn{1}{c|}{49.5$\downarrow$ }       & 61.9$\downarrow$         \\
\multicolumn{1}{c|}{}                                                                          & \multicolumn{1}{c|}{Down Stair}   & \multicolumn{1}{c|}{DTS}   & \multicolumn{1}{c|}{36.5$\downarrow$ }     & \multicolumn{1}{c|}{48.5$\downarrow$ }       & 60.9$\downarrow$         \\
\multicolumn{1}{c|}{}                                                                          & \multicolumn{1}{c|}{Bumpy Road}   & \multicolumn{1}{c|}{BM}    & \multicolumn{1}{c|}{41.3}     & \multicolumn{1}{c|}{53.3}       & 65.3        \\
\multicolumn{1}{c|}{}                                                                          & \multicolumn{1}{c|}{Curved Road}  & \multicolumn{1}{c|}{CV}    & \multicolumn{1}{c|}{46.2}     & \multicolumn{1}{c|}{58.5}       & 70.1        \\
\multicolumn{1}{c|}{}                                                                          & \multicolumn{1}{c|}{Soft Road}    & \multicolumn{1}{c|}{SF}    & \multicolumn{1}{c|}{42.5}     & \multicolumn{1}{c|}{53.7}       & 65.6        \\ \cmidrule{2-6}
\multicolumn{1}{c|}{}                                                                          & \multicolumn{1}{c|}{Average}      & \multicolumn{1}{c|}{-}     & \multicolumn{1}{c|}{39.3}     & \multicolumn{1}{c|}{51.1}       & \textbf{63.2}        \\ \hline \midrule

\multicolumn{1}{c|}{\multirow{4}{*}{Speed}}                                                    & \multicolumn{1}{c|}{Normal 1}     & \multicolumn{1}{c|}{NM1}   & \multicolumn{1}{c|}{43.9}     & \multicolumn{1}{c|}{55.2}       & 66.9        \\
\multicolumn{1}{c|}{}                                                                          & \multicolumn{1}{c|}{Fast}         & \multicolumn{1}{c|}{FA}    & \multicolumn{1}{c|}{32.8}     & \multicolumn{1}{c|}{42.9}       & 55.2        \\
\multicolumn{1}{c|}{}                                                                          & \multicolumn{1}{c|}{Stationary}   & \multicolumn{1}{c|}{ST}    & \multicolumn{1}{c|}{19.2$\downarrow$ }     & \multicolumn{1}{c|}{27.9$\downarrow$ }       & 40.9$\downarrow$         \\ \cmidrule{2-6}
\multicolumn{1}{c|}{}                                                                          & \multicolumn{1}{c|}{Average}      & \multicolumn{1}{c|}{-}     & \multicolumn{1}{c|}{32.0}     & \multicolumn{1}{c|}{42.0}       & \textbf{54.4}        \\ \hline \midrule

\multicolumn{1}{c|}{\multirow{5}{*}{\begin{tabular}[c]{@{}c@{}}Walking \\ Style\end{tabular}}} & \multicolumn{1}{c|}{Normal 2}     & \multicolumn{1}{c|}{NM2}   & \multicolumn{1}{c|}{43.9}     & \multicolumn{1}{c|}{55.3}       & 67.1        \\
\multicolumn{1}{c|}{}                                                                          & \multicolumn{1}{c|}{Confident}    & \multicolumn{1}{c|}{CF}    & \multicolumn{1}{c|}{42.0}     & \multicolumn{1}{c|}{54.2}       & 64.8        \\
\multicolumn{1}{c|}{}                                                                          & \multicolumn{1}{c|}{Freedom}      & \multicolumn{1}{c|}{FD}    & \multicolumn{1}{c|}{38.1}     & \multicolumn{1}{c|}{50.3}       & 63.0        \\
\multicolumn{1}{c|}{}                                                                          & \multicolumn{1}{c|}{Multi-Person} & \multicolumn{1}{c|}{MP}    & \multicolumn{1}{c|}{0.8$\downarrow$ }      & \multicolumn{1}{c|}{0.8$\downarrow$ }        & 1.0$\downarrow$          \\ \cmidrule{2-6}
\multicolumn{1}{c|}{}                                                                          & \multicolumn{1}{c|}{Average}      & \multicolumn{1}{c|}{-}     & \multicolumn{1}{c|}{31.2}     & \multicolumn{1}{c|}{40.2}       & \textbf{49.0}        \\ \bottomrule
\end{tabular}}
\label{Tab-3S}
\end{table}

\begin{table}[t!]
          \centering
          \caption{\textbf{Mixed-Covariate Evaluation}: R-1$^{hard}$ accuracy (\%) with excluding identical-view cases. 
          We use ``-" to connect the mixed covariates. Tab.~\ref{Tab-3} presents the dictionary containing abbreviations and their corresponding full spellings of these covariates. 
          $\downarrow$ and \textbf{bold} respectively indicate the sub-\textit{Average} and state-of-the-art performance.
          \footnotesize}
          \resizebox{1\columnwidth}{!}{
\begin{tabular}{ccccc}
\toprule
\multicolumn{5}{c}{Hard Metric}                                                                                                                                                                                          \\ \midrule
\multicolumn{2}{c|}{Publication}                                                                                                  & \multicolumn{1}{c|}{CVPR23}   & \multicolumn{1}{c|}{Arxiv23}    & Ours        \\ \midrule
\multicolumn{1}{c|}{Category}                                                                 & \multicolumn{1}{c|}{Covariate}    & \multicolumn{1}{c|}{GaitBase} & \multicolumn{1}{c|}{DeepGaitV2} & ParsingGait \\ \hline \midrule
\multicolumn{1}{c|}{\multirow{12}{*}{\begin{tabular}[c]{@{}c@{}}Two \\ Mixed\end{tabular}}}   & \multicolumn{1}{c|}{CL-UB}        & \multicolumn{1}{c|}{20.3$\downarrow$ }     & \multicolumn{1}{c|}{31.9$\downarrow$ }       & 40.4$\downarrow$         \\
\multicolumn{1}{c|}{}                                                                         & \multicolumn{1}{c|}{HVBX-BG}      & \multicolumn{1}{c|}{36.4}     & \multicolumn{1}{c|}{47.9}       & 60.5        \\
\multicolumn{1}{c|}{}                                                                         & \multicolumn{1}{c|}{BG-TC}        & \multicolumn{1}{c|}{38.8}     & \multicolumn{1}{c|}{50.4}       & 63.9        \\
\multicolumn{1}{c|}{}                                                                         & \multicolumn{1}{c|}{SF-CL}        & \multicolumn{1}{c|}{30.4$\downarrow$ }     & \multicolumn{1}{c|}{41.6$\downarrow$ }       & 53.8$\downarrow$         \\
\multicolumn{1}{c|}{}                                                                         & \multicolumn{1}{c|}{UTR-BX}       & \multicolumn{1}{c|}{36.4}     & \multicolumn{1}{c|}{47.5}       & 60.2        \\
\multicolumn{1}{c|}{}                                                                         & \multicolumn{1}{c|}{DTR-BK}       & \multicolumn{1}{c|}{37.8}     & \multicolumn{1}{c|}{49.1}       & 60.8        \\
\multicolumn{1}{c|}{}                                                                         & \multicolumn{1}{c|}{DTS-HVBX}     & \multicolumn{1}{c|}{31.7$\downarrow$ }     & \multicolumn{1}{c|}{43.7$\downarrow$ }       & 56.5$\downarrow$         \\
\multicolumn{1}{c|}{}                                                                         & \multicolumn{1}{c|}{UTS-BG}       & \multicolumn{1}{c|}{32.0$\downarrow$ }     & \multicolumn{1}{c|}{44.5$\downarrow$ }       & 58.4        \\
\multicolumn{1}{c|}{}                                                                         & \multicolumn{1}{c|}{BM-CL}        & \multicolumn{1}{c|}{28.9$\downarrow$ }     & \multicolumn{1}{c|}{41.0$\downarrow$ }       & 53.7$\downarrow$         \\
\multicolumn{1}{c|}{}                                                                         & \multicolumn{1}{c|}{CV-HVBX}      & \multicolumn{1}{c|}{42.6}     & \multicolumn{1}{c|}{54.9}       & 67.3        \\
\multicolumn{1}{c|}{}                                                                         & \multicolumn{1}{c|}{CL-CF}        & \multicolumn{1}{c|}{31.5$\downarrow$ }     & \multicolumn{1}{c|}{43.7$\downarrow$ }       & 55.7$\downarrow$         \\ \cmidrule{2-5}
\multicolumn{1}{c|}{}                                                                         & \multicolumn{1}{c|}{Average}      & \multicolumn{1}{c|}{33.4}     & \multicolumn{1}{c|}{45.1}       & \textbf{57.4}        \\ \hline \midrule

\multicolumn{1}{c|}{\multirow{12}{*}{\begin{tabular}[c]{@{}c@{}}Three \\ Mixed\end{tabular}}} & \multicolumn{1}{c|}{CL-UB-BG}     & \multicolumn{1}{c|}{19.4$\downarrow$ }     & \multicolumn{1}{c|}{30.6$\downarrow$ }       & 39.0$\downarrow$         \\
\multicolumn{1}{c|}{}                                                                         & \multicolumn{1}{c|}{BX-BG-CL}     & \multicolumn{1}{c|}{28.1}     & \multicolumn{1}{c|}{40.2}       & 51.3        \\
\multicolumn{1}{c|}{}                                                                         & \multicolumn{1}{c|}{BG-TC-CL}     & \multicolumn{1}{c|}{28.4}     & \multicolumn{1}{c|}{40.5}       & 53.6        \\
\multicolumn{1}{c|}{}                                                                         & \multicolumn{1}{c|}{SF-UB-BG}     & \multicolumn{1}{c|}{24.7$\downarrow$ }     & \multicolumn{1}{c|}{36.2$\downarrow$ }       & 47.4$\downarrow$         \\
\multicolumn{1}{c|}{}                                                                         & \multicolumn{1}{c|}{UTR-HVBX-CL}  & \multicolumn{1}{c|}{27.2$\downarrow$ }     & \multicolumn{1}{c|}{38.3$\downarrow$ }       & 49.5$\downarrow$         \\
\multicolumn{1}{c|}{}                                                                         & \multicolumn{1}{c|}{DTR-BK-BG}    & \multicolumn{1}{c|}{35.2}     & \multicolumn{1}{c|}{46.7}       & 58.9        \\
\multicolumn{1}{c|}{}                                                                         & \multicolumn{1}{c|}{DTS-HVBX-CL}  & \multicolumn{1}{c|}{23.4$\downarrow$ }     & \multicolumn{1}{c|}{34.4$\downarrow$ }       & 44.9$\downarrow$         \\
\multicolumn{1}{c|}{}                                                                         & \multicolumn{1}{c|}{UTS-BG-CL}    & \multicolumn{1}{c|}{23.1$\downarrow$ }     & \multicolumn{1}{c|}{34.4$\downarrow$ }       & 47.5$\downarrow$         \\
\multicolumn{1}{c|}{}                                                                         & \multicolumn{1}{c|}{BM-CL-BG}     & \multicolumn{1}{c|}{28.4}     & \multicolumn{1}{c|}{40.2}       & 52.5        \\
\multicolumn{1}{c|}{}                                                                         & \multicolumn{1}{c|}{CV-BX-BG}     & \multicolumn{1}{c|}{41.4}     & \multicolumn{1}{c|}{53.6}       & 65.9        \\
\multicolumn{1}{c|}{}                                                                         & \multicolumn{1}{c|}{UB-BG-FA}     & \multicolumn{1}{c|}{20.9$\downarrow$ }     & \multicolumn{1}{c|}{30.3$\downarrow$ }       & 40.5$\downarrow$         \\ \cmidrule{2-5}
\multicolumn{1}{c|}{}                                                                         & \multicolumn{1}{c|}{Average}      & \multicolumn{1}{c|}{27.3}     & \multicolumn{1}{c|}{38.7}       & \textbf{50.1}        \\ \hline \midrule

\multicolumn{1}{c|}{\multirow{9}{*}{\begin{tabular}[c]{@{}c@{}}Four \\ Mixed\end{tabular}}}   & \multicolumn{1}{c|}{CL-UB-BG-FA}  & \multicolumn{1}{c|}{15.9$\downarrow$ }     & \multicolumn{1}{c|}{24.7$\downarrow$ }       & 32.0$\downarrow$         \\
\multicolumn{1}{c|}{}                                                                         & \multicolumn{1}{c|}{BM-CL-BG-BX}  & \multicolumn{1}{c|}{27.5}     & \multicolumn{1}{c|}{39.2}       & 50.3        \\
\multicolumn{1}{c|}{}                                                                         & \multicolumn{1}{c|}{BG-TC-CL-CV}  & \multicolumn{1}{c|}{31.6}     & \multicolumn{1}{c|}{44.2}       & 57.9        \\
\multicolumn{1}{c|}{}                                                                         & \multicolumn{1}{c|}{DTR-BK-BG-CL} & \multicolumn{1}{c|}{27.3}     & \multicolumn{1}{c|}{38.6}       & 49.7        \\
\multicolumn{1}{c|}{}                                                                         & \multicolumn{1}{c|}{DTS-BX-CL-BG} & \multicolumn{1}{c|}{23.2}     & \multicolumn{1}{c|}{34.4}       & 45.1        \\
\multicolumn{1}{c|}{}                                                                         & \multicolumn{1}{c|}{SF-UB-BG-CL}  & \multicolumn{1}{c|}{18.4$\downarrow$ }     & \multicolumn{1}{c|}{28.7$\downarrow$ }       & 37.6$\downarrow$         \\
\multicolumn{1}{c|}{}                                                                         & \multicolumn{1}{c|}{BG-TC-CL-ST}  & \multicolumn{1}{c|}{10.9$\downarrow$ }     & \multicolumn{1}{c|}{16.5$\downarrow$ }       & 25.7$\downarrow$         \\
\multicolumn{1}{c|}{}                                                                         & \multicolumn{1}{c|}{UTS-UB-BG-CL} & \multicolumn{1}{c|}{15.1$\downarrow$ }     & \multicolumn{1}{c|}{24.6$\downarrow$ }       & 34.1$\downarrow$         \\ \cmidrule{2-5}
\multicolumn{1}{c|}{}                                                                         & \multicolumn{1}{c|}{Average}      & \multicolumn{1}{c|}{21.2}     & \multicolumn{1}{c|}{31.4}       & \textbf{41.5}        \\ \hline \midrule

\multicolumn{1}{c|}{\multirow{2}{*}{\begin{tabular}[c]{@{}c@{}}Five \\ Mixed\end{tabular}}}   & \multicolumn{1}{c|}{\begin{tabular}[c]{@{}c@{}}BG-TC-CL-\\ CV-UB\end{tabular}}  & \multicolumn{1}{c|}{20.1}     & \multicolumn{1}{c|}{31.5}       & \textbf{41.2}        \\ \cline{2-5} 
\multicolumn{1}{c|}{}                                                                         & \multicolumn{1}{c|}{\begin{tabular}[c]{@{}c@{}}UTR-BG-CL-\\ BX-CV\end{tabular}} & \multicolumn{1}{c|}{29.0}     & \multicolumn{1}{c|}{40.9}       & \textbf{52.3}        \\ \bottomrule
\end{tabular}}
\label{Tab-4S}
\end{table}

\bigskip

\bibliography{aaai24}

\begin{thebibliography}{46}
\providecommand{\natexlab}[1]{#1}

\bibitem[{{Altab Hossain} et~al.(2010){Altab Hossain}, Makihara, Wang, and
  Yagi}]{99}
{Altab Hossain}, M.; Makihara, Y.; Wang, J.; and Yagi, Y. 2010.
\newblock Clothing-invariant gait identification using part-based clothing
  categorization and adaptive weight control.
\newblock \emph{PR}, 43(6): 2281--2291.

\bibitem[{An et~al.(2020)An, Yu, Makihara, Wu, Xu, Yu, Liao, and Yagi}]{107}
An, W.; Yu, S.; Makihara, Y.; Wu, X.; Xu, C.; Yu, Y.; Liao, R.; and Yagi, Y.
  2020.
\newblock Performance Evaluation of Model-based Gait on Multi-view Very Large
  Population Database with Pose Sequences.
\newblock \emph{IEEE Trans. on Biometrics, Behavior, and Identity Science}.

\bibitem[{Cao et~al.(2017)Cao, Simon, Wei, and Sheikh}]{63}
Cao, Z.; Simon, T.; Wei, S.-E.; and Sheikh, Y. 2017.
\newblock Realtime Multi-Person 2D Pose Estimation Using Part Affinity Fields.
\newblock In \emph{CVPR}.

\bibitem[{Chao et~al.(2019)Chao, He, Zhang, and Feng}]{9}
Chao, H.; He, Y.; Zhang, J.; and Feng, J. 2019.
\newblock GaitSet: Regarding Gait as a Set for Cross-View Gait Recognition.
\newblock In \emph{AAAI}.

\bibitem[{Chen et~al.(2019)Chen, Pang, Wang, Xiong, Li, Sun, Feng, Liu, Shi,
  Ouyang, Loy, and Lin}]{109}
Chen, K.; Pang, J.; Wang, J.; Xiong, Y.; Li, X.; Sun, S.; Feng, W.; Liu, Z.;
  Shi, J.; Ouyang, W.; Loy, C.~C.; and Lin, D. 2019.
\newblock Hybrid task cascade for instance segmentation.
\newblock In \emph{CVPR}.

\bibitem[{Choi, Napolean, and van Gemert(2021)}]{121}
Choi, Y.; Napolean, Y.; and van Gemert, J.~C. 2021.
\newblock The Arm-Swing is Discriminative in Video Gait Recognition for Athlete
  Re-Identification.
\newblock In \emph{ICIP}.

\bibitem[{Ding et~al.(2022)Ding, Zhao, Liu, Zhang, and Peng}]{119}
Ding, T.; Zhao, Q.; Liu, F.; Zhang, H.; and Peng, P. 2022.
\newblock A Dataset and Method for Gait Recognition with Unmanned Aerial
  Vehicless.
\newblock In \emph{ICME}.

\bibitem[{Fan et~al.(2023{\natexlab{a}})Fan, Hou, Huang, and Yu}]{126}
Fan, C.; Hou, S.; Huang, Y.; and Yu, S. 2023{\natexlab{a}}.
\newblock Exploring Deep Models for Practical Gait Recognition.
\newblock \emph{ArXiv}, abs/2303.03301.

\bibitem[{Fan et~al.(2023{\natexlab{b}})Fan, Liang, Shen, Hou, Huang, and
  Yu}]{124}
Fan, C.; Liang, J.; Shen, C.; Hou, S.; Huang, Y.; and Yu, S.
  2023{\natexlab{b}}.
\newblock OpenGait: Revisiting Gait Recognition Towards Better Practicality.
\newblock In \emph{CVPR}, 9707--9716.

\bibitem[{Fan et~al.(2020)Fan, Peng, Cao, Liu, Hou, Chi, Huang, Li, and
  He}]{10}
Fan, C.; Peng, Y.; Cao, C.; Liu, X.; Hou, S.; Chi, J.; Huang, Y.; Li, Q.; and
  He, Z. 2020.
\newblock GaitPart: Temporal Part-Based Model for Gait Recognition.
\newblock In \emph{CVPR}.

\bibitem[{Fang et~al.(2017)Fang, Xie, Tai, and Lu}]{64}
Fang, H.-S.; Xie, S.; Tai, Y.-W.; and Lu, C. 2017.
\newblock RMPE: Regional Multi-Person Pose Estimation.
\newblock In \emph{ICCV}.

\bibitem[{Gong et~al.(2018)Gong, Liang, Li, Chen, Yang, and Lin}]{80}
Gong, K.; Liang, X.; Li, Y.; Chen, Y.; Yang, M.; and Lin, L. 2018.
\newblock Instance-Level Human Parsing via Part Grouping Network.
\newblock In \emph{ECCV}, 805--822.
\newblock ISBN 978-3-030-01225-0.

\bibitem[{Gross and Shi(2001)}]{93}
Gross, R.; and Shi, J. 2001.
\newblock The CMU Motion of Body (MoBo) Database.
\newblock \emph{Monumenta Nipponica}.

\bibitem[{Hofmann et~al.(2014)Hofmann, Geiger, Bachmann, Schuller, and
  Rigoll}]{103}
Hofmann, M.; Geiger, J.; Bachmann, S.; Schuller, B.; and Rigoll, G. 2014.
\newblock The TUM Gait from Audio, Image and Depth (GAID) database: Multimodal
  recognition of subjects and traits.
\newblock \emph{JVCIR}, 25(1): 195--206.

\bibitem[{Huang et~al.(2021)Huang, Zhu, Wang, Wang, Yang, He, Liu, and
  Feng}]{66}
Huang, X.; Zhu, D.; Wang, H.; Wang, X.; Yang, B.; He, B.; Liu, W.; and Feng, B.
  2021.
\newblock Context-Sensitive Temporal Feature Learning for Gait Recognition.
\newblock In \emph{ICCV}, 12909--12918.

\bibitem[{Iwama et~al.(2012)Iwama, Okumura, Makihara, and Yagi}]{101}
Iwama, H.; Okumura, M.; Makihara, Y.; and Yagi, Y. 2012.
\newblock The OU-ISIR Gait Database Comprising the Large Population Dataset and
  Performance Evaluation of Gait Recognition.
\newblock \emph{IEEE Trans. on Information Forensics and Security}, 7, Issue 5:
  1511--1521.

\bibitem[{Kirillov et~al.(2019)Kirillov, Girshick, He, and Dollar}]{114}
Kirillov, A.; Girshick, R.; He, K.; and Dollar, P. 2019.
\newblock Panoptic Feature Pyramid Networks.
\newblock In \emph{CVPR}.

\bibitem[{Li et~al.(2023)Li, Hou, Zhang, Cao, Liu, Huang, and Zhao}]{129}
Li, W.; Hou, S.; Zhang, C.; Cao, C.; Liu, X.; Huang, Y.; and Zhao, Y. 2023.
\newblock An In-Depth Exploration of Person Re-Identification and Gait
  Recognition in Cloth-Changing Conditions.
\newblock In \emph{CVPR}, 13824--13833.

\bibitem[{Li et~al.(2022)Li, Makihara, Xu, and Yagi}]{108}
Li, X.; Makihara, Y.; Xu, C.; and Yagi, Y. 2022.
\newblock Multi-View Large Population Gait Database With Human Meshes and Its
  Performance Evaluation.
\newblock \emph{IEEE Transactions on Biometrics, Behavior, and Identity
  Science}, 4(2): 234--248.

\bibitem[{Liang et~al.(2018)Liang, Gong, Shen, and Lin}]{78}
Liang, X.; Gong, K.; Shen, X.; and Lin, L. 2018.
\newblock Look into Person: Joint Body Parsing \& Pose Estimation Network and a
  New Benchmark.
\newblock \emph{IEEE TPAMI}.

\bibitem[{Lin, Zhang, and Yu(2021)}]{68}
Lin, B.; Zhang, S.; and Yu, X. 2021.
\newblock Gait Recognition via Effective Global-Local Feature Representation
  and Local Temporal Aggregation.
\newblock In \emph{ICCV}, 14648--14656.

\bibitem[{Liu et~al.(2004)Liu, Malave, Osuntogun, Sudhakar, and Sarkar}]{120}
Liu, Z.; Malave, L.; Osuntogun, A.; Sudhakar, P.; and Sarkar, S. 2004.
\newblock {Toward understanding the limits of gait recognition}.
\newblock In \emph{Biometric Technology for Human Identification}. SPIE.

\bibitem[{Makihara, Mannami, and Yagi(2011)}]{100}
Makihara, Y.; Mannami, H.; and Yagi, Y. 2011.
\newblock Gait Analysis of Gender and Age Using a Large-Scale Multi-view Gait
  Database.
\newblock In Kimmel, R.; Klette, R.; and Sugimoto, A., eds., \emph{ACCV},
  440--451. Berlin, Heidelberg: Springer Berlin Heidelberg.
\newblock ISBN 978-3-642-19309-5.

\bibitem[{Mansur et~al.(2014)Mansur, Makihara, Aqmar, and Yagi}]{98}
Mansur, A.; Makihara, Y.; Aqmar, R.; and Yagi, Y. 2014.
\newblock Gait Recognition under Speed Transition.
\newblock In \emph{CVPR}.

\bibitem[{Mu et~al.(2021)Mu, Castro, Mar\'in-Jim\'enez, Guil, ran Li, and
  Yu}]{77}
Mu, Z.; Castro, F.~M.; Mar\'in-Jim\'enez, M.~J.; Guil, N.; ran Li, Y.; and Yu,
  S. 2021.
\newblock {ReSGait}: The Real-Scene Gait Dataset.
\newblock In \emph{IJCB 2021}.

\bibitem[{Sarkar et~al.(2005)Sarkar, Phillips, Liu, Vega, Grother, and
  Bowyer}]{96}
Sarkar, S.; Phillips, P.; Liu, Z.; Vega, I.; Grother, P.; and Bowyer, K. 2005.
\newblock The humanID gait challenge problem: data sets, performance, and
  analysis.
\newblock \emph{IEEE TPAMI}, 27(2): 162--177.

\bibitem[{Shen et~al.(2023)Shen, Fan, Wu, Wang, Huang, and Yu}]{128}
Shen, C.; Fan, C.; Wu, W.; Wang, R.; Huang, G.~Q.; and Yu, S. 2023.
\newblock LidarGait: Benchmarking 3D Gait Recognition With Point Clouds.
\newblock In \emph{CVPR}, 1054--1063.

\bibitem[{{Shiraga} et~al.(2016){Shiraga}, {Makihara}, {Muramatsu}, {Echigo},
  and {Yagi}}]{17}
{Shiraga}, K.; {Makihara}, Y.; {Muramatsu}, D.; {Echigo}, T.; and {Yagi}, Y.
  2016.
\newblock GEINet: View-invariant gait recognition using a convolutional neural
  network.
\newblock In \emph{ICB}, 1--8.

\bibitem[{Shutler et~al.(2004)Shutler, Grant, Nixon, and Carter}]{95}
Shutler, J.~D.; Grant, M.~G.; Nixon, M.~S.; and Carter, J.~N. 2004.
\newblock On a large sequence-based human gaitdatabase.
\newblock In \emph{Applications and Science in Soft Computing}.

\bibitem[{Song et~al.(2022)Song, Huang, Wang, and Wang}]{127}
Song, C.; Huang, Y.; Wang, W.; and Wang, L. 2022.
\newblock CASIA-E: a large comprehensive dataset for gait recognition.
\newblock \emph{IEEE Transactions on Pattern Analysis and Machine
  Intelligence}, 45(3): 2801--2815.

\bibitem[{Sun et~al.(2019)Sun, Xiao, Liu, and Wang}]{111}
Sun, K.; Xiao, B.; Liu, D.; and Wang, J. 2019.
\newblock Deep High-Resolution Representation Learning for Human Pose
  Estimation.
\newblock In \emph{CVPR}.

\bibitem[{Takemura et~al.(2018)Takemura, Makihara, Muramatsu, Echigo, and
  Yagi}]{26}
Takemura, N.; Makihara, Y.; Muramatsu, D.; Echigo, T.; and Yagi, Y. 2018.
\newblock Multi-view large population gait dataset and its performance
  evaluation for cross-view gait recognition.
\newblock \emph{IPSJ Transactions on Computer Vision and Applications}, 10.

\bibitem[{Tan et~al.(2006)Tan, Huang, Yu, and Tan}]{97}
Tan, D.; Huang, K.; Yu, S.; and Tan, T. 2006.
\newblock Efficient Night Gait Recognition Based on Template Matching.
\newblock In \emph{ICPR)}, volume~3, 1000--1003.

\bibitem[{Teepe et~al.(2022)Teepe, Gilg, Herzog, H\"ormann, and Rigoll}]{122}
Teepe, T.; Gilg, J.; Herzog, F.; H\"ormann, S.; and Rigoll, G. 2022.
\newblock Towards a Deeper Understanding of Skeleton-Based Gait Recognition.
\newblock In \emph{CVPRW}.

\bibitem[{Teepe et~al.(2021)Teepe, Khan, Gilg, Herzog, Hörmann, and
  Rigoll}]{32}
Teepe, T.; Khan, A.; Gilg, J.; Herzog, F.; Hörmann, S.; and Rigoll, G. 2021.
\newblock Gaitgraph: Graph Convolutional Network for Skeleton-Based Gait
  Recognition.
\newblock In \emph{2021 IEEE International Conference on Image Processing
  (ICIP)}, 2314--2318.

\bibitem[{Uddin et~al.(2018)Uddin, Ngo, Makihara, Takemura, Li, Muramatsu, and
  Yagi}]{105}
Uddin, M.~Z.; Ngo, T.~T.; Makihara, Y.; Takemura, N.; Li, X.; Muramatsu, D.;
  and Yagi, Y. 2018.
\newblock The OU-ISIR Large Population Gait Database with real-life carried
  object and its performance evaluation.
\newblock \emph{IPSJ Transactions on Computer Vision and Applications}, 10(1):
  5.

\bibitem[{Wang et~al.(2019)Wang, Sun, Cheng, Jiang, Deng, Zhao, Liu, Mu, Tan,
  Wang, Liu, and Xiao}]{112}
Wang, J.; Sun, K.; Cheng, T.; Jiang, B.; Deng, C.; Zhao, Y.; Liu, D.; Mu, Y.;
  Tan, M.; Wang, X.; Liu, W.; and Xiao, B. 2019.
\newblock Deep High-Resolution Representation Learning for Visual Recognition.
\newblock \emph{IEEE TPAMI}.

\bibitem[{Wang et~al.(2003)Wang, Tan, Ning, and Hu}]{115}
Wang, L.; Tan, T.; Ning, H.; and Hu, W. 2003.
\newblock Silhouette analysis-based gait recognition for human identification.
\newblock \emph{IEEE TPAMI}.

\bibitem[{Xia et~al.(2017)Xia, Wang, Chen, and Yuille}]{81}
Xia, F.; Wang, P.; Chen, X.; and Yuille, A.~L. 2017.
\newblock Joint Multi-Person Pose Estimation and Semantic Part Segmentation.
\newblock In \emph{CVPR}.

\bibitem[{Xie et~al.(2021)Xie, Wang, Yu, Anandkumar, Alvarez, and Luo}]{110}
Xie, E.; Wang, W.; Yu, Z.; Anandkumar, A.; Alvarez, J.~M.; and Luo, P. 2021.
\newblock SegFormer: Simple and Efficient Design for Semantic Segmentation with
  Transformers.
\newblock In \emph{Advances in Neural Information Processing Systems}.

\bibitem[{Xu et~al.(2017)Xu, Makihara, Ogi, Li, Yagi, and Lu}]{104}
Xu, C.; Makihara, Y.; Ogi, G.; Li, X.; Yagi, Y.; and Lu, J. 2017.
\newblock The OU-ISIR Gait Database Comprising the Large Population Dataset
  with Age and Performance Evaluation of Age Estimation.
\newblock \emph{IPSJ Trans. on Computer Vision and Applications}, 9(24): 1--14.

\bibitem[{Yang et~al.(2021)Yang, Song, Wang, Liu, Xu, and Li}]{87}
Yang, L.; Song, Q.; Wang, Z.; Liu, Z.; Xu, S.; and Li, Z. 2021.
\newblock Quality-Aware Network for Human Parsing.
\newblock In \emph{arXiv preprint arXiv:2103.05997}.

\bibitem[{Yu, Tan, and Tan(2006)}]{25}
Yu, S.; Tan, D.; and Tan, T. 2006.
\newblock A Framework for Evaluating the Effect of View Angle, Clothing and
  Carrying Condition on Gait Recognition.
\newblock In \emph{ICPR}, volume~4, 441--444.

\bibitem[{Zhao et~al.(2018)Zhao, Li, Cheng, Sim, Yan, and Feng}]{79}
Zhao, J.; Li, J.; Cheng, Y.; Sim, T.; Yan, S.; and Feng, J. 2018.
\newblock Understanding Humans in Crowded Scenes: Deep Nested Adversarial
  Learning and A New Benchmark for Multi-Human Parsing.
\newblock In \emph{ACM MM}, 792–800.
\newblock ISBN 9781450356657.

\bibitem[{Zheng et~al.(2022)Zheng, Liu, Liu, He, Yan, and Mei}]{75}
Zheng, J.; Liu, X.; Liu, W.; He, L.; Yan, C.; and Mei, T. 2022.
\newblock Gait Recognition in the Wild With Dense 3D Representations and a
  Benchmark.
\newblock In \emph{CVPR}, 20228--20237.

\bibitem[{Zhu et~al.(2021)Zhu, Guo, Yang, Huang, Deng, Huang, Du, Lu, and
  Zhou}]{76}
Zhu, Z.; Guo, X.; Yang, T.; Huang, J.; Deng, J.; Huang, G.; Du, D.; Lu, J.; and
  Zhou, J. 2021.
\newblock Gait Recognition in the Wild: A Benchmark.
\newblock In \emph{ICCV}, 14789--14799.

\end{thebibliography}
	
\end{document}